\documentclass[journal]{IEEEtran}

\usepackage{amsmath,amssymb,amsfonts}
\usepackage{algorithmic}
\usepackage{graphicx}
\usepackage{textcomp}
 \usepackage[most]{tcolorbox}   
\usepackage{xcolor}
\usepackage[T1]{fontenc}
\usepackage[utf8]{inputenc}
\usepackage{array}
\usepackage{makecell}
\usepackage{multirow}
\usepackage{booktabs}
\usepackage{arydshln}
\usepackage[normalem]{ulem}
\usepackage{subcaption}
\usepackage{float}
\usepackage{svg}
\usepackage{tikz}
\usepackage{url}
\usepackage{stfloats}  
\usepackage{enumitem}
\usepackage{cite}          
\usepackage[hidelinks]{hyperref}   
\usepackage{tabularx}

\newcommand{\recscoredwat}[2]{\makecell{\footnotesize $\Delta$WA=#1\\\footnotesize $\Delta$BI=#2}}

\definecolor{grpref}{RGB}{90,90,90}     
\definecolor{grpsyn}{RGB}{31,94,166}    
\definecolor{obssyn}{RGB}{130,175,220}    

\definecolor{grpreal}{RGB}{200,90,20}   
\definecolor{obsreal}{RGB}{245,185,125}   

\definecolor{calib}{RGB}{120,180,120}   

\tcbset{recgroup/.style={
	colframe=#1, colback=#1!4, coltitle=white, colbacktitle=#1,
	arc=3pt, boxrule=0.8pt, toptitle=1pt, bottomtitle=1pt,
	left=1pt, right=1pt, top=2pt, bottom=1pt,
	fonttitle=\footnotesize\bfseries, halign title=center,
	nobeforeafter, box align=top}}

\newcommand{\recimg}[1]{\includegraphics[width=0.08\textwidth]{#1}}
\newcommand{\recscore}[2]{\makecell{\footnotesize PSNR=#1\\\footnotesize SSIM=#2}}

\newcommand{\recscorewat}[2]{\makecell{\footnotesize WA=#1\\\footnotesize BI=#2}}
\newcommand{\recvoid}{\vphantom{\recscore{0}{0}}} 

\renewcommand{\IEEEkeywordsname}{Keywords}

\colorlet{cbest}{blue!55!black}   
\colorlet{csec}{blue!60}    
\newcommand{\bstbest}[1]{\textcolor{csec}{\textbf{#1}}}
\newcommand{\secbest}[1]{\textcolor{cbest}{\textbf{#1}}}

\newenvironment{resume}%
  {\renewcommand{\abstractname}{R\'esum\'e}\begin{abstract}}%
  {\end{abstract}}
  
\usepackage{graphicx}
\usepackage{xcolor}       
\usepackage{import}
\usepackage{transparent}  

\title{Model-based neural network for global floodplains analysis}

\author{Pierre~Audisio$^{1}$,~Barbara~Belletti$^{2}$,~and~Nelly~Pustelnik$^{1}$\\[4pt]
{\normalsize\itshape
$^{1}$ENS de Lyon, CNRS, Laboratoire de Physique, F-69342 Lyon, France\\
$^{2}$Universit\'e Jean Monnet, CNRS, UMR 5600 Environnement Ville Soci\'et\'e, Saint-Etienne, France}\\[2pt]
}

\begin{document}

\IEEEaftertitletext{%
\vspace{-0.5\baselineskip}
\begin{resume}
Nous nous intéressons à la super-résolution d'images multispectrales acquises par télédétection, pour la surveillance à long terme des plaines inondables, en améliorant la résolution des acquisitions Landsat dans le but d'atteindre la même qualité de données que le programme Sentinel-2. 
En tirant parti de la période durant laquelle les deux missions se recouvrent temporellement, nous avons constitué $\mathcal{D}_{\text{Lhasa-L2S}}$, à notre connaissance le premier ensemble de données appariées dédié à la super-résolution multispectrale pour l’analyse fluviale. 
Il couvre la rivière Lhassa (Tibet), dont les fortes modifications anthropiques et la diversité des profils hydromorphologiques en font une zone d’étude idéale. 
Nous avons développé deux réseaux neuronaux basés sur des modèles : un schéma plug-and-play forward-backward (FB) et un schéma FB déroulé. 
Nous les avons comparés à des réseaux neuronaux de référence (SwinIR, RDN) sur des données synthétiques, sur des scènes réelles et à l’aide d’un test statistique porté sur l’ensemble du jeu de données réel apparié Landsat-Sentinel. 
Le réseau déroulé  atteint les valeurs les plus élevées de PSNR et de SSIM sur les composantes RGB réelles avec une inférence rapide, offrant le meilleur compromis entre précision et coût de calcul. 
Pour la segmentation de l’eau basée sur le NDWI, ses performances ne présentent aucune différence statistiquement significative par rapport à celles des meilleures méthodes.
\end{resume}

\begin{abstract}
We address the super-resolution of multispectral remote sensing imagery for long-term floodplain monitoring, enhancing Landsat acquisitions towards Sentinel-2 resolution. Exploiting the period during which both missions observed the same areas, we build $\mathcal{D}_{\text{Lhasa-L2S}}$, to our knowledge the first paired dataset dedicated to multispectral super-resolution for river analysis. It covers the Lhasa River (Tibet), whose strong anthropogenic changes and diverse hydromorphological profiles make it an ideal study area. We developed two model-based neural networks: a forward-backward (FB) plug-and-play scheme and an unfolded FB scheme. We compared them against learned neural networks baselines (SwinIR, RDN) on synthetic data, on real scenes, and through a statistical test over the whole real paired Landsat-Sentinel dataset. The unfolded network achieves the highest PSNR and SSIM on the real RGB components with fast inference, offering the best trade-off between accuracy and computational cost. For NDWI-based water segmentation, its performance shows no statistically significant difference from that of the best methods.
\end{abstract}

\vspace{1\baselineskip}}

\maketitle

\section{Introduction}
\label{sec:context}

\IEEEPARstart{I}{mage} restoration is a long-standing area of research that has seen significant progress over the past two decades, driven by developments in optimization \cite{Pustelnik_N_20016_j-w-enc-eee_wav_bid,Bauschke_H_2011_book_con_amo}, and the rise of neural networks \cite{ronneberger2015u}.
But despite groundbreaking progress in image recovery, many image restoration problems remain extremely challenging, at least for two main reasons. 
First, while state-of-the-art techniques perform well to provide ``good looking'' images for entertainment applications based on widely-used datasets, such as natural image collections or face datasets, {performance drops for images with  domain-specific structures, as those encountered in remote sensing imagery for scientific studies}  (e.g., climate, environment) and second, {difficulties persist in accurately recovering fine scale structures} when part of the image has complex local dynamics.

These two difficulties are jointly exemplified by the long-term study of extended floodplains at large scales, which relies on historical multispectral remote sensing imagery, notably in remote areas. Such images often suffer from low resolution, posing a significant challenge for long-term analysis.

But the incredible wealth of data opens many perspectives for reconstructing and understanding past environments and the changes they have undergone.
To achieve reliable monitoring over several decades, it becomes essential to increase the resolution of past low-resolution data  with high confidence in the recovery.

\noindent\textbf{Specificity of global floodplains analysis --}
Global floodplains cover less than 3\% of the Earth's surface, yet the richness of the ecosystem services they provide~\cite{schindler2014multifunctionality} makes them particularly attractive for population settlement. 
Over the past decades, human presence in floodplains has consequently increased in many parts of the world~\cite{fruh2015flood,han2020growth}. 
But climate change, together with the pressure of urbanization in these areas, constitutes a significant threat to both the local biodiversity and the health and safety of the population, given the increased flood risk associated with land-use changes~\cite{mazzoleni2021floodplains}.
To grasp these dynamics and risks, it is essential to analyze the evolution of river systems and their interactions with human activities over the long term \cite{schindler2014multifunctionality,fruh2015flood,han2020growth,mazzoleni2021floodplains,Crutzen2002}.

Doing so requires imagery that combines high spatial resolution, large spatial coverage, and a high temporal revisit frequency sustained over decades.
Satellite imagery meets several of these criteria, but no single mission meets all of them. 

\begin{figure}[h]
\includegraphics[width=\columnwidth]{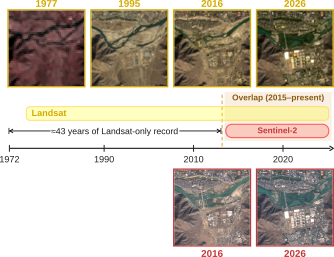}
\caption{Temporal coverage of data acquired by the Landsat and Copernicus programmes.}
\label{fig:used_data}
\end{figure}

Among freely available archives, few span periods as long as the Landsat program, which began in the 1970s~\cite{gascon2014}.
However, its spatial resolution (30\,m/px) is too coarse to resolve fine-scale change in detail.
Conversely, more recent missions such as Sentinel-2 (10\,m/px) from the Copernicus program~\cite{wulder2019} offer the required resolution but only since 2015, providing insufficient temporal depth for multi-decadal analysis (cf. Figure~\ref{fig:used_data}).
This study is motivated by resolving that trade-off: reconstructing high-resolution data with extensive temporal coverage through advanced single-image super-resolution, in order to upscale Landsat imagery from 30\,m to 10\,m and thereby approach Sentinel-2 quality across the entire historical archive.

\noindent\textbf{Single image super-resolution (SISR)} --  
Single image super-resolution is a well known problem \cite{irani1991,freeman2000learning}  that has been solved for many years considering model-based variational approaches including the use of total variation (TV) penalization~\cite{babacan2010variational} or penalization based on wavelet transform \cite{ji2008robust}.
However, despite the effectiveness of these classical techniques, the ability of data-based (i.e. deep learning) to achieve outstanding restoration quality has driven a shift towards neural networks approaches, where model parameters are now learned directly from data  \cite{nazzal2015}. While hybrid model/data-based methods have proven a strong compromise between purely model-based and purely data-driven approaches in standard RGB image restoration, this direction remains comparatively underexplored in remote sensing SISR, despite its potential to improve interpretability, performance, and robustness.\\ 

\noindent\textbf{SISR overview focused on remote sensing} --   In the specific case of remote sensing multispectral data acquisition, single image super-resolution has attracted growing interest  (see \cite{li2021singleimage,rasti2022image} for an overview), and has been extensively studied to overcome the spatial-resolution limit of Earth-observation sensors \cite{fernandez2017single,esr2022,constantin2021joint}. 
%
Beyond single-band optical data, dedicated efforts target the joint spatial and spectral structure of multi- and hyperspectral images, whose reconstruction must preserve inter-band coherence \cite{tsagkatakis2019,han2025recent,loncan2015hyperspectral}, supported by benchmark spectral scenes and datasets \cite{arad2016sparse,chakrabarti2011statistics,yasuma2010generalized,yi2020aerial,hu2023mdas}.

Early SISR methods relied on frequency-domain signal processing (Fourier, wavelet transforms) \cite{demirel2011discrete}, which were computationally efficient but struggled with complex structure recovery. Then, interpolation-based methods such as bicubic, bilinear or nearest-neighbour were developed but produce blurred edges and artifacts from lost high-frequency detail \cite{mareboyana2016high}. The next step was involving reconstruction-based methods relying on variational approaches with hand-designed priors, including non-local sparse penalization \cite{gou2014remote, chierchia2014nonlocal}, to sharpen edges and reduce artifacts, sometimes combined with fast multilevel procedures to handle the large dimensionality of the images \cite{lauga2022methodes}. 

Beyond this reliance on explicit degradation models \cite{fernandez2017single}, the field has since shifted toward learning-based approaches, starting with convolutional neural networks (CNNs) \cite{lei2017superresolution}, residual dense network (RDN) \cite{zhang2018residual},  and, more recently, generative adversarial networks (GANs)
\cite{Tuna2018,wang2021lightweight,ren2021remote,zhang2020unsupervised,Kim2020,Prajapati2020} (cf. \cite{esr2022} for a detailed review focused on deep learning).
The more efficient methods using GANs need to make very strong assumptions about the distribution of data for training the prior, which is not always known on remote sensing images, and this can lead to a degradation of the results and the creation of artifacts \cite{daniels2020reducing}. 
Supervised approaches also involve models other than CNNs, such as Transformers \cite{vaswani2017attention}.
The term “Transformer” refers to the self-attention mechanism, which models the global relationships among image regions, eliminating the reliance on content-independent convolution kernels used in CNNs.
One of the most promising methods in image super-resolution using this approach is SwinIR \cite{liang2021swinir}. 

Finally, model-based neural network approaches, including unfolded networks and Plug-and-Play (PnP) methods offer a compelling alternative, combining the interpretability and robustness of model-based variational methods with the expressiveness of data-driven approaches. PnP methods rely on splitting algorithms such as forward-backward or alternating direction method of multipliers (ADMM) \cite{Venkatakrishnan2013,park2025plug,hurault2024convergent} to decouple the action of the data-fidelity and prior terms, allowing a learned CNN to be used as a denoiser in place of a hand-designed proximal operator \cite{moreau1965}.

In remote sensing, model-based deep networks have only recently begun to emerge, notably through unfolded architectures 
\cite{wang2022,han2025recent,lin2026deep}, but they remain scarce and have yet to be applied to SISR. 
PnP methods are similarly underexplored in this domain: \cite{tao2020} and \cite{lai2022deep} report performance competitive with state-of-the-art neural networks, yet with notable limitations in scope. 
The former focuses on hyperspectral images without a specific application context, while the latter presents a preliminary study restricted to RGB imagery from the U.S. Geological Survey National Map Urban Area collection.

\noindent\textbf{Contribution} -- We cast the Landsat-to-Sentinel reconstruction problem, with a focus on floodplain analysis, as a specific scientific question. In this context, we deeply investigate model-based neural network approaches, namely Plug-and-Play and unfolded frameworks, by instantiating both on multispectral remote sensing data.
Concretely, our contribution is fourfold: (i) we build dedicated datasets tailored to this reconstruction task; (ii) we adapt two state-of-the-art deep denoisers, a UNet~\cite{ronneberger2015u} and a DRUNet~\cite{zhang2021}, to 4-channel multispectral inputs, and deploy them both as pretrained PnP priors and as learned components of an unfolded architecture; (iii) we benchmark this family of physics-aware reconstructions, involving the forward operator, against two purely data-driven baselines, RDN~\cite{zhang2018residual} and SwinIR~\cite{liang2021swinir}; (iv) we analyze the quality of the reconstruction with respect to the scientific objective.

\section{Problem Formulation}
\label{sec:forward_model}
The key element distinguishing a model-based neural network from pure neural network is the incorporation of the forward model into the architecture. Model-based neural networks combine this forward operator with a denoiser within an algorithmic scheme, enforcing data consistency while encoding prior knowledge. 

In this section, we start by describing the forward model and the FB building block of both PnP and unfolded schemes. Then, we specify the PnP and unfolded architectures for the specific purpose of the Landsat-to-Sentinel reconstruction problem, along with their respective training procedures. We then describe the state-of-the-art fully data-driven NN methods used for comparison and summarize the pros and cons of each approach.

\subsection{Physics-based reconstructions design}
\noindent\textbf{Forward model --} Following~\cite{wang2020deep,wei2015fast}, the forward model relating the Landsat (LR) and Sentinel (HR) images is defined for each band $b\in\{1,\ldots,B\}$ as
\begin{equation}
\mathrm{z}_b = (\mathrm{A}\,\bar{\mathrm{x}}_b)\!\downarrow_s + \varepsilon,
\label{eq:forward}
\end{equation}
where $\bar{\mathrm{x}} = (\bar{\mathrm{x}}_b)_{1\leq b\leq B}\in\mathbb{R}^{BN}$ is the original HR image (Sentinel-2), $\mathrm{A}$ denotes the blurring operator, $\downarrow_s$ the downsampling operator with decimation factor $s=3$, $\varepsilon\sim\mathcal{N}(\mathbf{0},\sigma^2\mathbf{I}_d)$ a white Gaussian noise, and $\mathrm{z}\in\mathbb{R}^{BM}$ the LR observation (Landsat-8/9). The calibration is further discussed in Sec.~\ref{sec:dataset}.

\noindent\textbf{Forward-backward building block --}
We explore the capacity of Plug-and-Play and unfolded approaches
to recover multispectral remote sensing images. Both strategies considered in this work are built on the forward-backward (proximal-gradient) algorithm~\cite{hurault2022proximal,pesquet2021learning}, whose iterations read
\begin{equation}
    \label{eq:fb}
    \mathrm{x}_{k+1} = \mathrm{D}_{\theta,\gamma,\lambda}\big(\mathrm{x}_k - \gamma\,\mathrm{A}^{\top}(\mathrm{A}\mathrm{x}_k - \mathrm{z})\big),
\end{equation}
where $\gamma>0$ is a step size and $\mathrm{D}_{\theta,\gamma,\lambda}$ a neural network that can be interpreted as a Gaussian denoiser.
PnP and unfolded approaches are then distinguished solely by how
$\mathrm{D}_{\theta,\gamma,\lambda}$ is trained, a distinction we detail in the
following two sections. 
To simplify the presentation, we introduce the notation
\begin{equation}
    \label{eq:unfolded}
    \mathrm{\widehat{x}} = \Phi_{\Theta}^{K}(\mathrm{z}),
    \quad \text{with} \quad
   \Phi_{\Theta}^{K} = \varphi_{\theta_K} \circ \cdots \circ \varphi_{\theta_1},
\end{equation}
and, for $k=1,\ldots,K$,
\begin{equation}
\label{eq:itFB}
    (\forall\, \mathrm{x} \in \mathbb{R}^N) \qquad
    \varphi_{\theta_k}(\mathrm{x}) = \mathrm{D}_{\theta_k,\gamma,\lambda}\big(\mathrm{x} - \gamma\,\mathrm{A}^{\top}(\mathrm{A}\mathrm{x} - \mathrm{z})\big).
\end{equation}
Each layer thus combines a linear step, enforcing consistency with the data-fidelity term, with a non-linear step, enforcing prior knowledge, performed by the denoiser $\mathrm{D}_{\theta_k,\gamma,\lambda}$; stacking $K\geq 1$ such layers forms the resulting network.

$\bullet$ \textit{Plug-and-Play (FB-PnP)} -- 
The denoiser $\mathrm{D}_{\theta,\gamma,\lambda}$ is trained once, offline, 
and simply reused, unchanged, inside the iteration~\eqref{eq:fb}.   %
%
In our notation, this amounts to setting
$\theta_1=\cdots=\theta_K=\widehat{\theta}$ in Eq.~\eqref{eq:unfolded}, where
\begin{equation}
    \label{eq:pnp_training}
    \widehat{\theta} \in \operatorname*{Argmin}_{\theta}
    \sum_{(\overline{\mathrm{x}}, \mathrm{y}) \in \mathcal{T}_{\text{PnP}}}
    \big\Vert \overline{\mathrm{x}} - \mathrm{D}_{\theta,\gamma,\lambda}(\mathrm{y}) \big\Vert^2
\end{equation}
is learned on a training set $\mathcal{T}_{\text{PnP}}$ of
clean HR images $\overline{\mathrm{x}}$ paired with their noisy counterparts
$\mathrm{y} = \overline{\mathrm{x}} + \eta$ with $\eta\sim\mathcal{N}(0, \tilde{\sigma} \mathbb{I})$ and lead to $\widehat{\Theta}_{\text{PnP}} = (\widehat{\theta},\ldots,\widehat{\theta})$. The resulting denoiser is then plugged into the PnP iteration
and run for $K$ iterations (typically of order 100 or more to achieve stability in PSNR evolution), yielding the reconstruction 
\begin{equation}
    \mathrm{\widehat{x}}_{\text{PnP}} = \Phi_{\widehat{\Theta}_{\text{PnP}}}^{K}(\mathrm{z}),
    \qquad
    \Phi_{\widehat{\Theta}_{\text{PnP}}}^{K} = \underbrace{\varphi_{\widehat{\theta}} \circ \cdots \circ \varphi_{\widehat{\theta}}}_{K \text{ times}}.
\end{equation}
Because the denoiser never sees the actual degradation operator $\mathrm{A}$
or the reconstruction objective during training, $(\widehat{\theta},\gamma,\lambda)$
are not jointly optimized, which requires manual tuning and limits the adaptability of the prior to the task-specific degradation.
Convergence can be recovered under structural constraints on it, non-expansiveness, gradient-step, or maximal monotonicity~\cite{pesquet2021learning,hurault2022proximal,hurault2024convergent}, with recent connections to score-based sampling~\cite{park2025plug}. While open implementations now make such solvers broadly available for RGB image restoration, comparable tools for multispectral images remain scarce~\cite{tachella2025deepinverse}.

Four our purpose, we thus instantiate the denoiser $\mathrm{D}_{\theta,\gamma,\lambda}$ either as a Unet or a DRUNet (a hybrid U-Net \cite{ronneberger2015u} and ResNet \cite{he2016deep} architecture), both adapted from the DeepInverse library \cite{tachella2025deepinverse} by modifying its first and last layers to handle 4-channel multispectral data.
It was trained on HR Sentinel-2 images corrupted by white Gaussian noise, for 1500 epochs using an AdamW optimiser (learning rate $10^{-3}$) with a StepLR scheduler (step size 100, decay factor 0.5), and a batch size of 4.

$\bullet$ \textit{Unfolded Networks (FB-Unfolded)} -- Algorithm unrolling~\cite{gregor2010learning,monga2021algorithm,jiu2021deep,bertocchi2020deep} overcomes the PnP limitation by jointly optimizing  $({\theta},\gamma,\lambda)$. Each
forward-backward step becomes a layer, and the resulting network is trained
end-to-end on the reconstruction task~\cite{shlezinger2023model}. 
Unlike PnP, the parameters $\theta_1,\ldots,\theta_K$ are no longer tied together nor learned independently of the degradation model, instead, the full set $\widehat{\Theta}_{\text{unf}}=(\widehat{\theta}_k,\widehat{\gamma}_k, \widehat{\lambda}_k)_{k\in\{1,\ldots,K\}}$ is learned jointly, directly from paired LR-HR data, as
\begin{equation}
    \label{eq:unfolded_training_1}
    \widehat{\Theta}_{\text{unf}} \in \operatorname*{Argmin}_{\Theta}
    \sum_{(\overline{\mathrm{x}}, \mathrm{z}) \in \mathcal{T}_{\text{full}}}
    \big\Vert \overline{\mathrm{x}} - \Phi_{\Theta}^{K}(\mathrm{z}) \big\Vert^2,
\end{equation}
on a training set $\mathcal{T}_{\text{full}}$ of clean HR images
$\overline{\mathrm{x}}$ paired with their corresponding LR observations
$\mathrm{z}$, obtained via the forward model of Eq.~\eqref{eq:forward}. The
resulting reconstruction is given by
\begin{equation}
    \mathrm{\widehat{x}}_{\text{unf}} = \Phi_{\widehat{\Theta}_{\text{unf}}}^{K}(\mathrm{z}).
\end{equation}
Because the forward operator $\mathrm{A}$ is embedded in every layer and the denoiser weights ${\theta}_k$ are allowed to differ across layers, the network can compensate for the specific degradation at hand. The number
of layers $K$ is typically much smaller than in PnP (of order 10). 
The joint learning of $({\theta},\gamma,\lambda)$ yields a
model-informed architecture with greater adaptability to the specific degradation, at the cost of requiring supervised paired data and a fixed
computational budget $K$.\\

\noindent\textbf{Theoretical properties --} The forward-backward iteration~\eqref{eq:fb} can formally be related either
to the minimization problem
\begin{equation}
    \label{eq:argmin}
    \widetilde{\mathrm{x}}_{\text{min}} \in \operatorname*{Argmin}_{\mathrm{x}}
  \tfrac{1}{2}\|\mathrm{A}\mathrm{x} - \mathrm{z}\|^2_2
    \;+\;
    g(\mathrm{x}),
\end{equation}
or to an inclusion problem of the form
\begin{equation}
    \label{eq:inclusion}
    0 \in \mathrm{A}^\top(\mathrm{A}\widetilde{\mathrm{x}}_{\text{inc}} - \mathrm{z}) + \mathbf{T}(\widetilde{\mathrm{x}}_{\text{inc}}).
\end{equation}
The link
with~\eqref{eq:argmin} requires $\mathrm{D}_{\theta,\gamma,\lambda}$ to be
related to the proximity operator of $g$~\cite{moreau1965}. Under this
correspondence, convergence of the sequence $(\mathrm{x}_k)_{k\in\mathbb{N}}$
to $\widetilde{\mathrm{x}}_{\text{min}}$ is guaranteed for specific choices of the step-size~\cite{combettes2005signal,gribonval2020characterization,hurault2022proximal}, depending on properties of $g$ (e.g., convexity). When
$\mathrm{D}_{\theta,\gamma,\lambda}$ cannot be interpreted as a proximity operator, convergence of $(\mathrm{x}_k)_{k\in\mathbb{N}}$ to a solution $\widetilde{\mathrm{x}}_{\text{inc}}$ of the inclusion problem~\eqref{eq:inclusion} is still guaranteed, provided that $\mathbf{T}$ is related to $\mathrm{D}_{\theta,\gamma,\lambda}$ and the denoiser is constrained to be firmly non-expansive during training~\cite{pesquet2021learning}.

In the PnP setting, since $\mathrm{D}_{\theta,\tau,\lambda}$ is fixed across
all $K$ iterations and $K$ can be taken arbitrarily large, the asymptotic
theory outlined above applies directly. In the unfolded setting, by
contrast, the network has a fixed, finite depth $K$,
and the layer-wise parameters $\theta_1,\ldots,\theta_K$ are allowed to
differ across layers. As a result, the asymptotic convergence guarantees
available for PnP no longer hold: the output $\Phi_{\widehat{\Theta}}^K(\mathrm{z})$
is not the limit of an infinite iterative process, but a truncated,
task-specific approximation of $\widehat{\mathrm{x}}$. Recovering stronger
guarantees in this setting requires turning to \emph{Deep Equilibrium} (DEQ)
formulations~\cite{crockett_bilevel_2022,gilton_deep_2021,davy2025restarted}.

\subsection{Data-driven baselines}
To benchmark the proposed PnP and unfolded reconstructions against
standard state-of-the-art architectures, we consider two networks
well established for super-resolution. Similarly as for unfolded, the training is 
\begin{equation}
    \label{eq:unfolded_training_2}
    \widehat{\Theta} \in \operatorname*{Arg\,min}_{\Theta}
    \sum_{(\overline{\mathrm{x}}, \mathrm{z}) \in \mathcal{T}_{\text{full}}}
    \big\Vert \overline{\mathrm{x}} - \Phi_{\Theta}(\mathrm{z}) \big\Vert^2.
\end{equation}
Here, $\Phi_{\Theta}$ is simply a stack of linear and activation layers, with no forward model $\mathrm{A}$ embedded in its architecture and no relation to \eqref{eq:unfolded}. We therefore drop the superscript $K$ to avoid confusion with the unfolded network. The training dataset,
however, remains the same as for the unfolded approach, i.e.
$\mathcal{T}_{\text{full}}$.

The first, the Residual Dense Network (RDN)~\cite{zhang2018residual}, is a purely convolutional, data-driven method reaching about $1.2\,\mathrm{M}$ of parameters in our configuration. 
RDN mainly consists of four parts: shallow feature extraction net, residual dense blocks (RDBs), dense feature fusion (DFF), and finally the up-sampling net (UPNet).
The second, SwinIR~\cite{liang2021swinir}, is a transformer-based model built on shifted-window self-attention, composed by three modules: shallow feature extraction, deep feature extraction and high-quality (HQ) image reconstruction modules. Shallow feature extraction use a convolutional layer, whereas deep feature extraction is performed by several residual Swin Transformer blocks (RSTB) Then for the final reconstruction shallow and deep features are aggregated.
With roughly $12\,\mathrm{M}$ parameters and represents the current attention-based state-of-the-art for single-image super-resolution.

Contrary to our model-based schemes, none of the two incorporates the forward operator explicitly: RDN and SwinIR learn the inverse mapping end-to-end as a black box. 


\subsection{Comparison between the architectures}
\label{sec:architectures}
The reconstruction techniques considered span a diverse range of methods, from "informed" approaches offering theoretical guarantees of varying strength, to purely neural-network-based approaches. Among the informed approaches, we consider architectures with varying numbers of weights. Table~\ref{tab:training} summarizes the information regarding all the compared strategies.

\section{Lhasa-L2S: A New Multispectral River Dataset}
\label{sec:dataset}

To perform super-resolution of Landsat data,  we train a super-resolution model calibrated on the contemporary overlap (from 2015 to present) of Landsat/Sentinel-2 programmes, with a long-term goal to apply it retrospectively to imagery reaching back to the 1970s.  The sampling is task-specific, with pairs centered on river corridors rather than drawn from a general-purpose, which concentrates the training distribution on the fluvial morphology.


\noindent\textbf{Necessity to create a new database --} 
Super-resolution of freely available multispectral imagery has attracted considerable attention, with numerous recent methods and dedicated paired datasets. Closely related to the present study, Wang et al.~\cite{wang2026benchmark} recently proposed the Land2Sent dataset of paired Landsat-8/9 and Sentinel-2 images. It consists of 30 image pairs acquired in 2023, of size $7700 \times 7800$ pixels for Landsat and $10980 \times 10980$ pixels for Sentinel-2, with acquisition time differences ranging from 14 to 23 minutes. Although valuable, this dataset is too restrictive in terms of acquisition time to allow for sufficient diversity in river samples crucial for the purpose of this study. More generally,  such  Landsat/Sentinel dataset  remains largely unexplored.
The dominant strategy is at the opposite to pair Sentinel-2 as the low-resolution source with a very-high-resolution reference. For instance, SatDiffMoE~\cite{satdiffmoe} builds on the fMoW archive~\cite{fmow}, whose imagery originates from QuickBird-2 and WorldView-3, launched in 2001 and 2014 respectively. 
WorldStrat~\cite{worldstrat} pairs SPOT-6/7 acquisitions with Sentinel-2. 
The benchmark of~\cite{kowaleczko} also takes Sentinel-2 as input, but this time with WorldView-2 as high-resolution reference, while SEN2NAIP~\cite{sen2naip} substitutes high-resolution with aerial imagery from the National Agriculture Imagery Program.  Another line of work pairs Sentinel-2 with commercial PlanetScope imagery~\cite{kopec}. 

\begin{table}[t]
	\centering
	\footnotesize
	\setlength{\tabcolsep}{4pt}
	\begin{tabular}{@{}l|l|c|l|r|r@{}}
		\textbf{Method} & \textbf{Architecture} & \textbf{Training} & \textbf{Data} & \textbf{Param.} & \textbf{Inf. time}\\
		\hline
		\multirow{2}{*}{FB-PnP}
		& DRUNet & $\mathcal{L}_{\mathbf{D}}$ & $\mathcal{T}=\mathcal{T}_{\text{PnP}}$ & 32\,M   & 4.55\\
		& UNet   & $\mathcal{L}_{\mathbf{D}}$ & $\mathcal{T}=\mathcal{T}_{\text{PnP}}$ & 8.5\,M  & 1.04\\
		\hline
		\multirow{2}{*}{FB-Unf.}
		& DRUNet & $\mathcal{L}_{\Phi}$ & $\mathcal{T}=\mathcal{T}_{\text{full}}$ & 652\,M   & 0.39\\
		& UNet   & $\mathcal{L}_{\Phi}$ & $\mathcal{T}=\mathcal{T}_{\text{full}}$ & 171\,M & 0.09\\
		\hline
		SwinIR & Transformers & $\mathcal{L}_{\Phi}$ & $\mathcal{T}=\mathcal{T}_{\text{full}}$ & 11.9\,M & 0.04\\
		\hline
		\multirow{2}{*}{RDN}
		& Residual     & \multirow{2}{*}{$\mathcal{L}_{\Phi}$} & \multirow{2}{*}{$\mathcal{T}=\mathcal{T}_{\text{full}}$} & \multirow{2}{*}{1.2\,M} & \multirow{2}{*}{0.003}\\
		& dense block  & & & &\\
		\hline
	\end{tabular}
	\caption{Training configuration of the plug-and-play, unfolded, RDN and SwinIR models.
	Two training objectives are used: the denoising loss
	$\mathcal{L}_{\mathbf{D}}=\underset{\Theta}{\min}\sum_{(\overline{\mathrm{x}},\mathrm{z})\in\mathcal{T}}\big\Vert\overline{\mathrm{x}}-\mathbf{D}_{\Theta}(\mathrm{z})\big\Vert^2$,
	which trains the denoiser alone, and the end-to-end reconstruction loss
	$\mathcal{L}_{\Phi}=\underset{\Theta}{\min}\sum_{(\overline{\mathrm{x}},\mathrm{z})\in\mathcal{T}}\big\Vert\overline{\mathrm{x}}-\Phi_{\Theta}(\mathrm{z})\big\Vert^2$.
	In parameters, M stands for millions; inference time is reported per $300\times300$ image on a single GPU. Inference time is given in seconds.}
	\label{tab:training}
\end{table}

Two limitations follow from this common design. First, the reference sensors are predominantly commercial. QuickBird, WorldView, SPOT-6/7, and PlanetScope all operate under restrictive licensing, which constrains the volume of training data that can be assembled, limits independent reproduction, and precludes open redistribution of the resulting pairs. Second, and more decisive for long-term monitoring, none of these platforms dates from before the 2000s. The historical depth of any dataset built on them is therefore bounded by the launch date of its reference sensor, and the resulting models cannot be applied to imagery acquired before that horizon. Harmonisation efforts such as HLS~\cite{claverie2018} do bring Landsat and Sentinel-2 into a common radiometric frame through consistent atmospheric correction, BRDF normalisation, spatial co-registration and band matching, but they operate at the coarser of the two resolutions and do not attempt reconstruction. Only a small number of recent studies~\cite{sambandham} address the Landsat-to-Sentinel direction explicitly, and none targets a specific geomorphological object.

Monitoring natural processes and those driven by human activity requires a database with a short revisit interval sustained over several years. Without it, the full evolution of a process cannot be interpreted through the extraction of specific indicators or through visual analysis by experts. Multi-source databases combining hyperspectral, multispectral, and SAR imagery do exist~\cite{yi2020aerial,hu2023mdas}, but they remain limited either in spatial extent or in temporal coverage. Studies spanning several years over large areas instead require a database that is consistent along both dimensions at once, dense in time and broad in space, so that spatial correlations can be fully accounted for and tracked over time. These considerations motivated the construction of a paired database of Sentinel-2~\cite{gascon2014} and Landsat-8/9~\cite{wulder2019} imagery for the calibration, training, and validation of super-resolution models. 
Both sources are freely available, which removes the licensing constraint described above. 
Sentinel-2 serves as the high-resolution reference at 10\,m with a 5-day revisit, against 30\,m and 8 days for the combined Landsat-8/9 constellation, and the shorter revisit permits the construction of high-fidelity pairs in which land-surface occupancy is nearly identical between the two sources. 

In this work we consider only four spectral bands, namely the red, green, blue, and near-infrared (NIR) bands. 
This choice is motivated by two reasons. 
First, these bands allow the computation of the main indices required for monitoring water bodies, such as the NDWI (see Section~\ref{sec:experiments}), and the surrounding natural environment. 
Second, they are the bands available at the finest spatial resolution for Sentinel-2 (10~m), and they share a common spatial resolution for Landsat 8 and 9 (30~m).

\begin{figure}[h]
\includegraphics[width=\columnwidth]{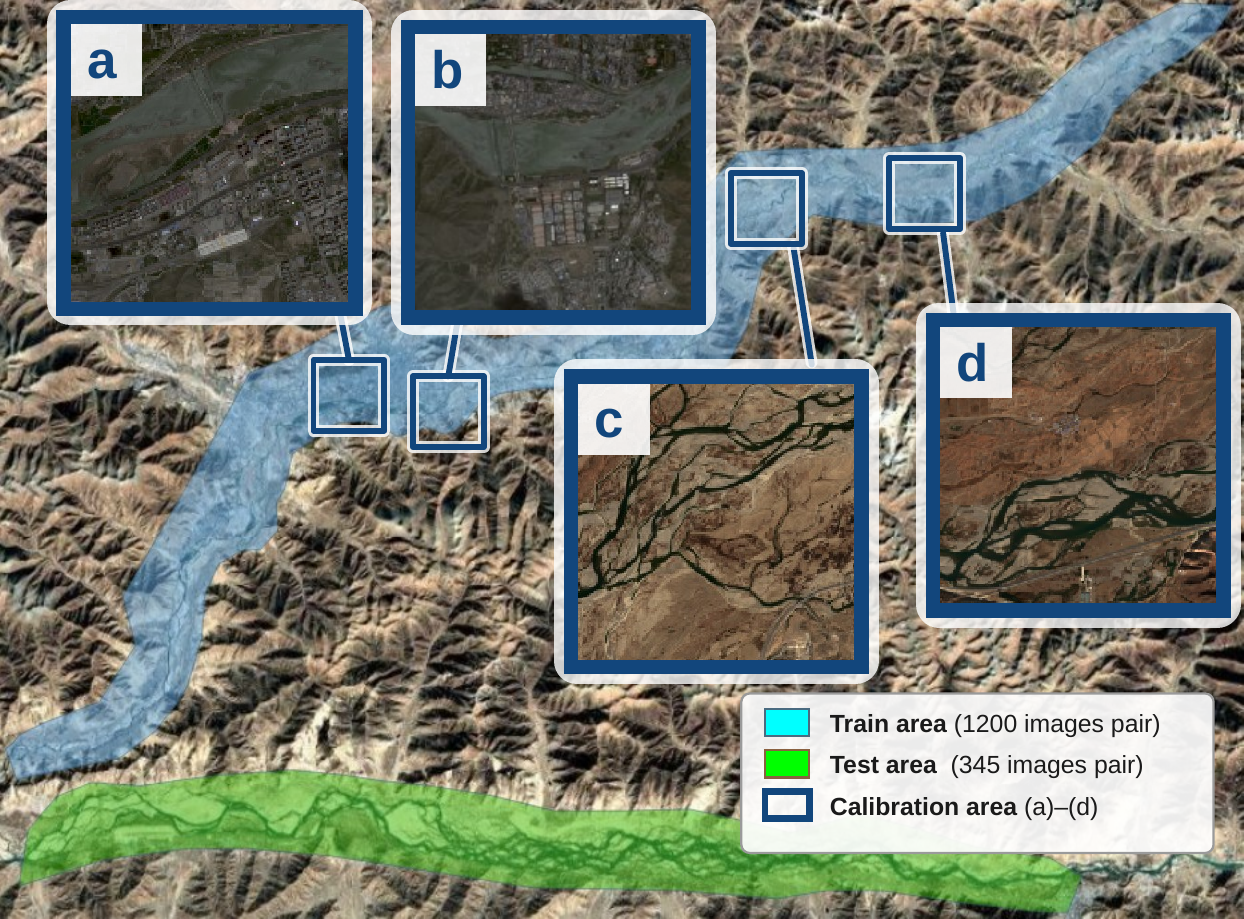}
\caption{Study area of the Lhasa River in Tibet, showing the locations of the areas used for training (blue area), validation (a,b,c,d) and the test phase (green area).}
\label{fig:test_train_area}
\end{figure}

\noindent\textbf{Paired-database construction --} Existing databases such as Land2Sent~\cite{wang2026benchmark} are not directly suited to river-specific applications, as described above. 
To address this gap, we build a new dataset, denoted $\mathcal{T}_{\text{Lhasa-L2S}}$, by adapting the pipeline originally used to construct Land2Sent, incorporating a more recent deep-learned cloud filter provided by Google Earth Engine (GEE) and restricting the image selection to a river-specific geographical extent. 
The corridor is partitioned geographically rather than randomly, so that the corridor crops used for training and those used for testing are spatially disjoint, corresponding respectively to the blue and green areas in Figure~\ref{fig:test_train_area}, a protocol motivated by the performance drop observed when remote sensing models are transferred across regions~\cite{nyborg2022timematch}. The final proposed dataset $\mathcal{T}_{\text{Lhasa-L2S}}$ is produced through the following steps:

\begin{enumerate}[label=(\roman*)]
	\item \textbf{Acquisition.} Images covering the Lhasa River (Figure~\ref{fig:test_train_area}) are downloaded through the GEE API~\cite{gorelick2017google}. These images are Level 2A products, which means they have undergone pre-processing to remove the optical effects caused by the Earth’s atmosphere.
	\item \textbf{Tile-level cloud screening.} Landsat-8/9 acquisitions whose scene cloud cover exceeds $30\%$ are discarded to exclude unusable images (notably during the rainy season, June--August), while each Sentinel-2 acquisition is masked with Cloud Score+~\cite{pasquarella2023}, a learned model that jointly detects low and high clouds and their shadows and assigns a per-pixel clarity score (band \texttt{cs\_cdf}), here thresholded at $0.65$. 
	\item \textbf{Study-area screening.} Only tiles whose fraction of valid pixels over the area of interest exceeds a minimum threshold are kept.  
    \item \textbf{Spatial pairing.} A Landsat/Sentinel-2 match is retained only if the two footprints share a common overlap of at least $100\times100$\,km$^2$ (i.e. $10^4\times 10^4$ px for Sentinel). 
	\item \textbf{Temporal pairing.} As the two sensors cannot acquire simultaneously, each Landsat acquisition is paired with a single Sentinel-2 image within a maximum temporal separation of $7$ days (in line with the revisit periods of Sentinel-2, 5 days, and Landsat-8/9, 8 days) and this window was empirically verified to ensure sufficient consistency of ground coverage.
	\item \textbf{Centerline cropping.} Using the spatial alignment, both sources are cropped along the centerline of the river corridor~\cite{pekel2016high} into image pairs $(\overline{\mathrm{x}},\mathrm{z})$, where $\overline{\mathrm{x}}$ denotes the high-resolution (HR) image ($300\,\text{px}\times300\,\text{px}$) and $\mathrm{z}$ the low-resolution (LR) image ($100\,\text{px}\times100\,\text{px}$).
	\item \textbf{Crop validation.} A crop is retained only if both images contain at least $99\%$ valid pixels and no more than $0.1\%$ of no-data or non-finite values. 
    A user-defined overlap between successive crops regulates redundancy along the corridor, in our case it was set to 30\%.
    \item \textbf{Tensor creation.} For each validated crop, a single tensor is created, stacking the four spectral bands of interest: red, green, blue, and NIR. This enabled the subsequent download of a single tensor containing all the necessary information, rather than retrieving the four bands as separate images.
	\item \textbf{Output.} The resulting HR/LR tensor pairs, together with their geographic coordinates, constitute the main dataset $\mathcal{D}_{\text{Lhasa}}$.  From an initial set of 85 full Landsat-8/9 scenes and 110 Sentinel-2 scenes over the blue area in Figure~\ref{fig:test_train_area}, 1212 image pairs were extracted at crop scale, reducing the dataset size from 104.8GB to  2.4GB. For the green area, 690 pairs are extracted.
\end{enumerate}

\noindent\textbf{Partition --} 
The pipeline yields to the database $\mathcal{D}_{\text{Lhasa-L2S}}$ composed with $1902$ HR/LR pairs over the blue area and green area of 
Figure~\ref{fig:test_train_area}, split as follows:
\begin{itemize}
    \item $1200$ pairs for training (from blue area)
    \item $8$ pairs for calibration (from blue area)
    \item $4$ pairs for validation  (from blue area)
    \item $690$ pairs for statistical test (from green area)
    
\end{itemize}
Beyond their number, these pairs were selected to cover a wide variety of landscapes, ranging from heavily urbanised areas, dominated by built-up areas and infrastructure, to predominantly rural environments comprising agricultural and natural areas with a low density of buildings, and including  many intermediate configurations that combine these two types of environment. 
This heterogeneity is deliberate, as exposing the model to a diversity of landscape yields a network that remains reliable over territories undergoing active urbanization and, more generally, over areas subject to anthropogenic pressure. 

In detail, the pairs are split into four disjoint sets with distinct roles.  The largest set, containing 1200 pairs, is used for training.   The four validation pairs, denoted (a),(b),(c),(d) and shown in Figure~\ref{fig:test_train_area}, were set aside prior to training, so that they remain independent of the data used to fit the models. 
They serve two purposes, first, they support the evaluation of the reconstructions, both quantitatively through the PSNR and visually, by inspecting the fidelity of the restored textures with respect to the Sentinel-2 reference. 
Second, they allow an extensive comparison between the different trained models, as well as the calibration of the hyper-parameters of the PnP approach, in particular the regularization parameter $\lambda$ and the step size $\gamma$. 
Since the four pairs cover a wide diversity of land uses (urban and rural areas), calibrating these parameters on them make the reconstruction robust to the different types of scene encountered in the dataset. The 8 calibration pairs are used during training, without gradient computation, to monitor the evolution of the PSNR across epochs, this makes it possible to follow the convergence of the reconstruction, to detect overfitting where it occurs, and to select the checkpoint used for inference before the PSNR starts to decrease at later epochs.
Finally, the 690 test pairs are used for the statistical comparison of the models, so that the differences in reconstruction quality can be assessed on a large test set (see Section~\ref{sec:stat}).

\begin{figure}[t]
  \centering
  \def\svgwidth{\columnwidth}
  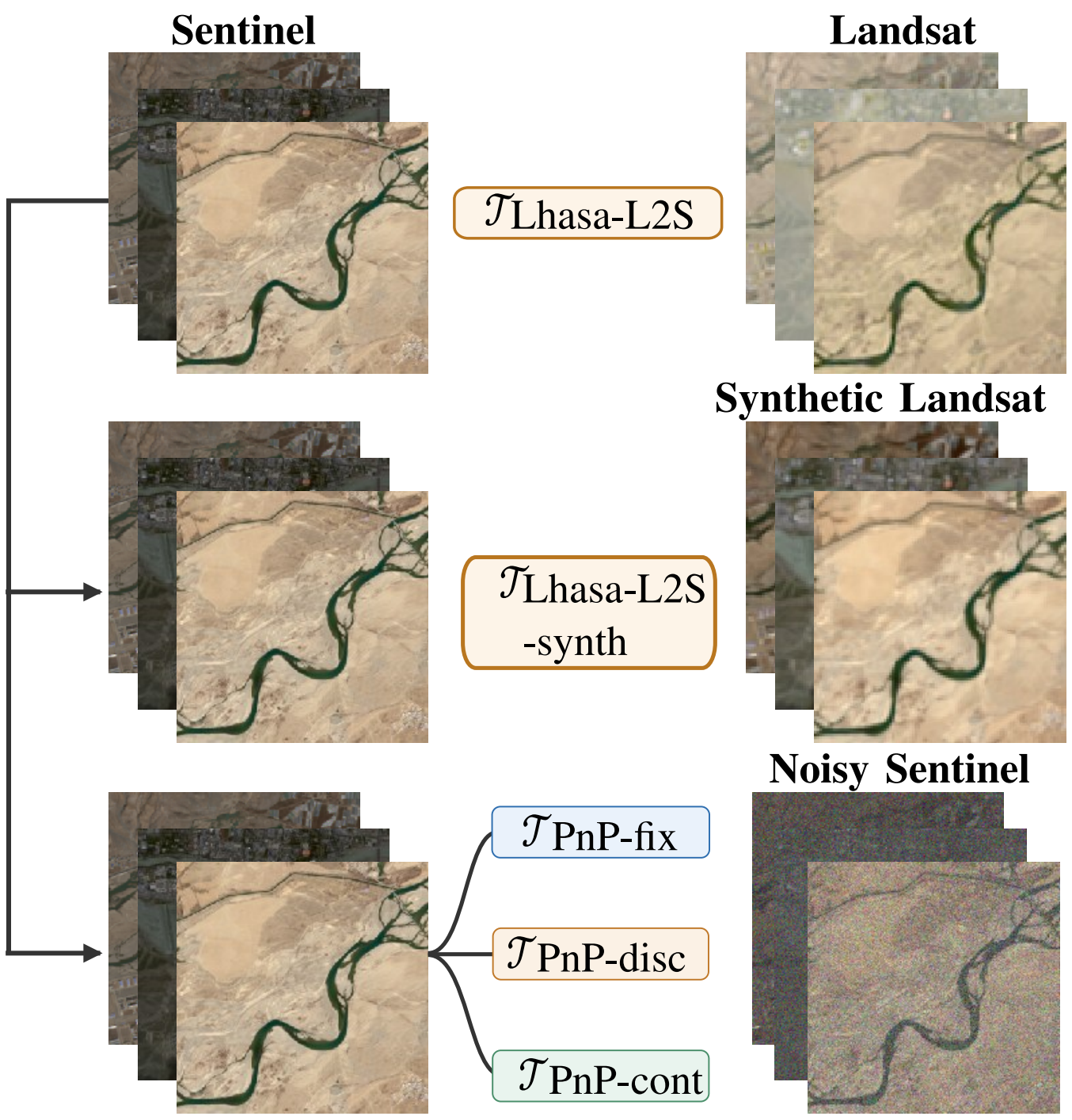
  \caption{Main database $\mathcal{T}_{\text{Lhasa-L2S}}$ and the corresponding training datasets: $\mathcal{T}_{\text{Lhasa-L2S-synth}}$ (blur $+$ downsampling $+$ noise) for the unfolded approach, and the three $\mathcal{T}_{\text{PnP}}$ training sub-databases for the PnP approach.}
  \label{fig:test_train}
\end{figure}

\noindent\textbf{Training sub-databases --} $\mathcal{T}_{\text{Lhasa-L2S}}$ dataset  refers to the training/calibration subset of $\mathcal{D}_{\text{Lhasa-L2S}}$. This dataset is central 
for building two derived datasets: $\mathcal{T}_{\text{Lhasa-L2S-synt}}$ synthetic dataset whose high-resolution images are taken from $\mathcal{T}_{\text{Lhasa-L2S}}$ and whose low-resolution counterparts are generated with the forward model of Eq.~\eqref{eq:forward}, and $\mathcal{T}_{\text{Lhasa-L2S-PnP}}$, required for training the denoiser in PnP strategy. These three datasets are illustrated in Figure~\ref{fig:test_train}. 

\noindent $\bullet$ \; \textit{$\mathcal{T}_{\text{Lhasa-L2S}}$ real dataset}\\
This dataset is composed of 1208 pairs of high- and low-resolution images, acquired respectively by the Sentinel-2 and Landsat-8/9 satellites, of size $300\times300$ (HR) and $100\times100$ (LR), see the \textit{Partition} sub-section: training + calibration.

\noindent $\bullet$ \; \textit{$\mathcal{T}_{\text{Lhasa-L2S-synth}}$ synthetic dataset}\\
This dataset is a synthetic counterpart of $\mathcal{T}_{\text{Lhasa-L2S}}$ composed of 1208 pairs.
Instead of using real observation from Landsat project as it is in $\mathcal{T}_{\text{Lhasa-L2S}}$, we create a low resolution image using the forward model of Equation~\eqref{eq:forward}, applied to the high resolution images.
This dataset makes it possible to calibrate the models in a simpler setting, free of the difficulties raised by real data. 
Indeed, on the real dataset the forward model only approximates the relation between the Landsat-8/9 observation and the Sentinel-2 reference, whereas on the synthetic dataset this relation holds exactly by construction, since the low-resolution image is produced from the reference itself. 
Indeed, in $\mathcal{T}_{\text{Lhasa-L2S}}$ the reference and observation images of each pair are not acquired simultaneously, and differences in viewing angle, illumination, atmospheric conditions, and river discharge may therefore be present between the two, which could negatively affect the performance of the models.

\noindent $\bullet$ \; \textit{$\mathcal{T}_{\text{Lhasa-L2S-PnP}}$ dataset for PnP method}\\ For the PnP approach, as mentioned earlier, a specific dataset has to be created.  The choice of the noise level used during training is crucial, as it can significantly affect the super-resolution results. We therefore consider several strategies, with either a fixed or a variable noise level, and build three sub-databases, each containing noisy versions of the high-resolution images generated according to different choices for $\widetilde{\sigma}$: $\mathcal{T}_{\text{PnP-fixed}}$, $\mathcal{T}_{\text{PnP-discrete}}$, and $\mathcal{T}_{\text{PnP-uniform}}$, in which the Gaussian noise level $\widetilde{\sigma}$ is either fixed, drawn from a discrete set, or drawn from a uniform distribution (see Figure~\ref{fig:sigma_tree}). The $\mathcal{T}_{\text{PnP}}$ sub-database contains 1212 pairs, each composed of a high-resolution image and a noisy version of it, both of size $300\times300$. We distinguish three $\mathcal{T}_{\text{PnP}}$ regimes:

\begin{IEEEdescription}[\IEEEsetlabelwidth{$\mathcal{T}_{\text{PnP-fixvar}}$: }]
\item[\normalfont$\mathcal{T}_{\text{PnP-fixed}}$:]
A denoiser is trained per noise level, for $\widetilde{\sigma} \in \{0.015, 0.03, 0.045, 0.06, 0.075, 0.1, 0.5, 0.9\}$.

\item[\normalfont$\mathcal{T}_{\text{PnP-disc.}}$:]
A single denoiser is trained, with the noise level drawn randomly from a discrete set $\widetilde{\sigma} \in \{0.045, 0.06, 0.075, 0.09, 0.1\}$.

\item[\normalfont$\mathcal{T}_{\text{PnP-cont.}}$:]
Three denoisers are trained, with the noise level drawn randomly either from a uniform distribution on $[0,1]$ or on $[0,0.2]$, or from a Gaussian distribution with mean $\mu = 0.015$ and standard deviation $0.04$.
\end{IEEEdescription}


\usetikzlibrary{decorations.pathreplacing,calc}

\definecolor{cfix}{HTML}{0072B2}
\definecolor{cdis}{HTML}{D55E00}
\definecolor{cuni}{HTML}{009E73}
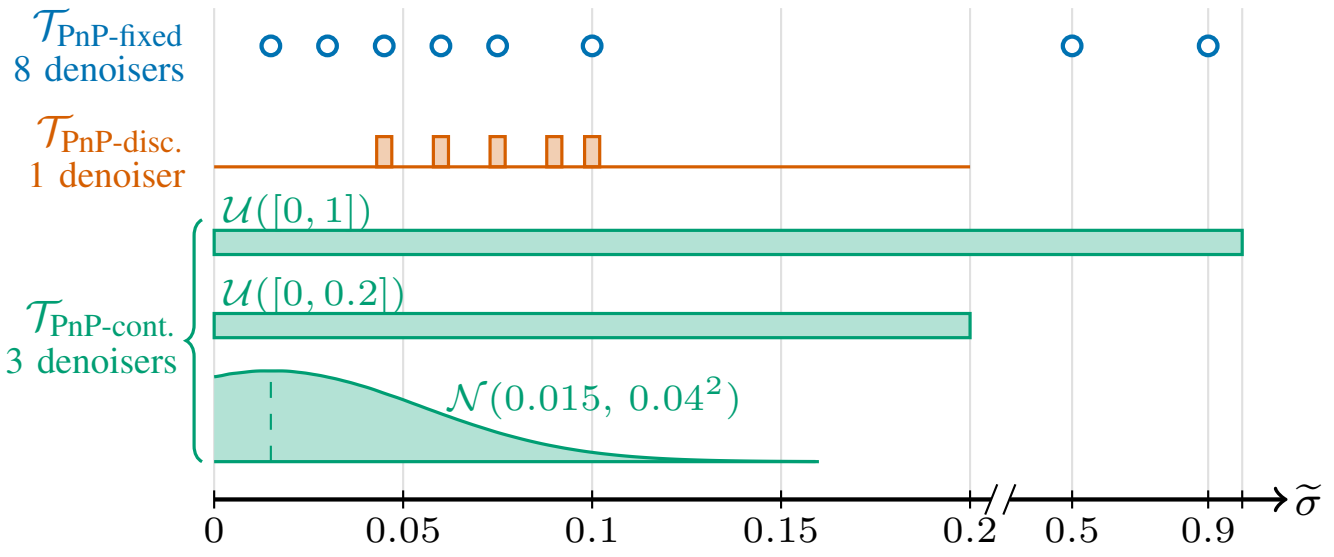
\begin{figure}[h]
\begin{tikzpicture}[
    font=\footnotesize,
    declare function={
      X(\s) = (\s <= 0.2) * (25*\s) + (\s > 0.2) * (5.0 + 2.25*(\s-0.2));
    },
    dot/.style={circle, inner sep=0pt, minimum size=3.5pt},
    lab/.style={left, align=right, inner sep=1pt},
  ]

  \foreach \s in {0, 0.05, 0.1, 0.15, 0.2, 0.5, 0.9, 1}
    \draw[black!12] ({X(\s)},0) -- ({X(\s)},3.25);

  \draw[->, thick] (0,0) -- ({X(1)+0.3},0) node[right, inner sep=1pt] {$\widetilde{\sigma}$};
  \foreach \s/\l in {0/0, 0.05/0.05, 0.1/0.1, 0.15/0.15, 0.2/0.2, 0.5/0.5, 0.9/0.9}
    \draw ({X(\s)},0.06) -- ({X(\s)},-0.06) node[below, inner sep=1.5pt] {\scriptsize $\l$};
  \draw ({X(1)},0.06) -- ({X(1)},-0.06);
  \fill[white] ({X(0.2)+0.14},-0.1) rectangle ({X(0.2)+0.26},0.1);
  \draw ({X(0.2)+0.10},-0.1) -- ({X(0.2)+0.18},0.1);
  \draw ({X(0.2)+0.22},-0.1) -- ({X(0.2)+0.30},0.1);

  \def\yfix{3.0}
  \node[lab, text=cfix] at (-0.15,\yfix)
    {$\mathcal{T}_{\text{PnP-fixed}}$\\[-1pt]{\scriptsize 8 denoisers}};
  \foreach \s in {0.015, 0.03, 0.045, 0.06, 0.075, 0.1, 0.5, 0.9}
    \node[dot, draw=cfix, fill=white, thick] at ({X(\s)},\yfix) {};

  \def\ydis{2.2}   
  \def\hdis{0.2}   
  \def\wdis{0.05}  
  \node[lab, text=cdis] at (-0.15,\ydis+0.1)
    {$\mathcal{T}_{\text{PnP-disc.}}$\\[-1pt]{\scriptsize 1 denoiser}};
  \foreach \s in {0.045, 0.06, 0.075, 0.09, 0.1} {
    \fill[cdis!30] ({X(\s)-\wdis},\ydis) rectangle ({X(\s)+\wdis},\ydis+\hdis);
    \draw[cdis, semithick] ({X(\s)-\wdis},\ydis) rectangle ({X(\s)+\wdis},\ydis+\hdis);
  }
  \draw[cdis, semithick] (0,\ydis) -- ({X(0.2)},\ydis);

  \def\yua{1.7}   
  \def\yub{1.15}  
  \def\yg{0.25}   
  \def\bh{0.16}   
  \def\gh{0.6}    

  \fill[cuni!30] (0,\yua-\bh/2) rectangle ({X(1)},\yua+\bh/2);
  \draw[cuni, semithick] (0,\yua-\bh/2) rectangle ({X(1)},\yua+\bh/2);
  \node[right, text=cuni, font=\scriptsize, inner sep=1pt] at (0.02,\yua+0.2) {$\mathcal{U}([0,1])$};

  \fill[cuni!30] (0,\yub-\bh/2) rectangle ({X(0.2)},\yub+\bh/2);
  \draw[cuni, semithick] (0,\yub-\bh/2) rectangle ({X(0.2)},\yub+\bh/2);
  \node[right, text=cuni, font=\scriptsize, inner sep=1pt] at (0.02,\yub+0.2) {$\mathcal{U}([0,0.2])$};

  \fill[cuni!30] (0,\yg)
    -- plot[domain=0:0.16, samples=80, smooth]
       ({X(\x)}, {\yg + \gh*exp(-((\x-0.015)^2)/(2*0.04^2))})
    -- ({X(0.16)},\yg) -- cycle;
  \draw[cuni, semithick]
    plot[domain=0:0.16, samples=80, smooth]
       ({X(\x)}, {\yg + \gh*exp(-((\x-0.015)^2)/(2*0.04^2))});
  \draw[cuni, semithick] (0,\yg) -- ({X(0.16)},\yg);
  \draw[cuni, dashed] ({X(0.015)},\yg) -- ({X(0.015)},\yg+\gh);
  \node[right, text=cuni, font=\scriptsize, inner sep=1pt] at ({X(0.06)},\yg+0.42)
    {$\mathcal{N}(0.015,\,0.04^2)$};

  \draw[decorate, decoration={brace, amplitude=3pt, mirror}, cuni, semithick]
    (-0.08,\yua+0.15) -- (-0.08,\yg);
  \node[lab, text=cuni] at (-0.2,{(\yua+\yg)/2+0.1})
    {$\mathcal{T}_{\text{PnP-cont.}}$\\[-1pt]{\scriptsize 3 denoisers}};

\end{tikzpicture}
	\caption{Training strategies.}
	\label{fig:sigma_tree}
\end{figure}

\noindent\textbf{Validation sub-databases --} All the methods (PnP, unfolded, SwinIR, and RDN) are evaluated on a validation dataset built in the same way as the training real datasets, which does not share  image with the training data. 

The real validation dataset is denoted $\mathcal{V}_{\text{Lhasa-L2S}}$ real dataset. It consists of 694 pairs of images, each pairing a high-resolution ($300\times300$) Sentinel-2 image with a low-resolution ($100\times100$) Landsat-8/9 image (see the \textit{Partition} subsection: validation + statistical test). The first set of 4 images enables a fine-grained, image-by-image analysis of the results. It is then extended with 694 additional images which provides a more exhaustive and statistically representative evaluation.
The synthetic counterpart validation dataset is denoted $\mathcal{V}_{\text{Lhasa-L2S-synt}}$.

\section{Numerical Experiments}
\label{sec:experiments}

Before presenting the reconstruction experiments, it is important to recall the objectives of this study. 
The goal is 
to correctly reconstruct the hydromorphology of the river, both in rural areas and in urbanised ones, in order to capture changes in the river structure: the anthropogenic modifications that develop over several years along and within the riverbanks, and the seasonal variations between the dry and the wet seasons. 
An accurate image reconstruction allows both short- and long-term monitoring of the impact of these changes on the natural dynamics of the river.

\subsection{Evaluation Criteria}
\label{sec:eval_criteria}
Reconstruction quality is evaluated with standard metrics, namely the peak signal-to-noise ratio (PSNR) and the structural similarity index (SSIM). These standard metrics are very useful for comparing our approach with other state-of-the-art methods, but they are agnostic to the hydromorphological information, which is ultimately the information to recover: two super-resolved images with comparable PSNR may still differ substantially in the river features they render. We therefore complement them with two domain-specific metrics computed on the reconstructed river corridor. These metrics rely on the Normalized Difference Water Index (NDWI)~\cite{gao1996ndwi}.
For an image $\mathrm{x} \in \mathbb{R}^{B \times N}$ with $B$ spectral bands and $N$ pixels, the NDWI is computed at each pixel $n \in \{1, \dots, N\}$ from the green and near-infrared (NIR) bands as
\begin{equation}
	\label{eq:ndwi}
	\mathrm{d}_n^{\mathrm{NDWI}} = \frac{\mathrm{x}_{\mathrm{green},n} - \mathrm{x}_{\mathrm{NIR},n}}{\mathrm{x}_{\mathrm{green},n} + \mathrm{x}_{\mathrm{NIR},n}}.
\end{equation} 
 The four metrics are defined below and rely on the reference image  of $N$ pixels $\bar{\mathrm{x}}$ and its reconstruction $\widehat{\mathrm{x}}$.

\begin{itemize}

\item \textbf{Peak signal-to-noise ratio (PSNR):} It is defined as
\begin{equation}
	\begin{split}
		\mathrm{PSNR}(\bar{\mathrm{x}},\widehat{\mathrm{x}}) &= 10 \log_{10} \left( \frac{\mathrm{max}(\widehat{\mathrm{x}})^{2}}{\Vert \bar{\mathrm{x}}-\widehat{\mathrm{x}}\Vert^2} \right). \\
	\end{split}
	\label{eq:psnr}
\end{equation}

\item \textbf{Structural similarity index mesure (SSIM):} It compares local statistics and is given by
\begin{equation}
	\mathrm{SSIM}(\bar{\mathrm{x}},\widehat{\mathrm{x}}) = \frac{\left( 2 \mu_{\bar{\mathrm{x}}} \mu_{\widehat{\mathrm{x}}} + c_{1} \right) \left( 2 \sigma_{\bar{\mathrm{x}}\widehat{\mathrm{x}}} + c_{2} \right)}{\left( \mu_{\bar{\mathrm{x}}}^{2} + \mu_{\widehat{\mathrm{x}}}^{2} + c_{1} \right) \left( \sigma_{\bar{\mathrm{x}}}^{2} + \sigma_{\widehat{\mathrm{x}}}^{2} + c_{2} \right)},
	\label{eq:ssim}
\end{equation}
where $\mu_{\bar{\mathrm{x}}}$ and $\mu_{\widehat{\mathrm{x}}}$ are the local means, $\sigma_{\bar{\mathrm{x}}}^{2}$ and $\sigma_{\widehat{\mathrm{x}}}^{2}$ the local variances, $\sigma_{\bar{\mathrm{x}}\widehat{\mathrm{x}}}$ the local covariance, and $c_{1} = (k_{1}\mathrm{max}(\widehat{\mathrm{x}}))^{2}$ and $c_{2} = (k_{2}\mathrm{max}(\widehat{\mathrm{x}}))^{2}$ two stabilising constants with $k_{1} = 0.01$ and $k_{2} = 0.03$~\cite{wang2004image}.

\item \textbf{Open-water surface area (WA):}  It is delineated through the NDWI thresholding, such as 

\begin{equation}
\mathrm{t}_n^{\mathrm{NDWI}} =
\begin{cases}
1 & \text{if } \mathrm{d}_n^{\mathrm{NDWI}} > \tau, \\
0 & \text{otherwise,}
\end{cases}
\end{equation}

with $\tau = 0.1$, in line with established surface-water extraction practice~\cite{pekel2016high} giving the following formula: $\textrm{WA} = \eta \sum \mathrm{t}_n^{\mathrm{NDWI}}$ expressed in $m2$ where $\eta = 10\times 10$. 
The water extent recovered from the reconstruction is then compared to that of the Sentinel-2 reference, measuring how faithfully the method restores the spectral dynamics. 

\item \textbf{Braiding index (BI):} It is a morphological descriptor characterizing the multi-thread planform of the river~\cite{egozi2008}. It is computed from  $\mathrm{t}^{\mathrm{NDWI}}$, from which water pixels (i.e. $\mathrm{t}_n^{\mathrm{NDWI}}\equiv 1$)  having no water pixel among their four edge-sharing neighbours (4-neighbour connectivity)  are first removed in order to suppress isolated misclassifications. The mask is then reduced by morphological thinning to a one-pixel-wide skeleton, which is converted into an undirected graph $\mathcal{G} = (V,E)$:   end points (one neighbour) and junctions (three or more) form the vertices $V$, and each chain of pixels linking two vertices forms an edge $e \in E$. Its length $L_{e}$  is obtained by summing the distances between successive pixels of the chain (a horizontal or vertical step contributing $1$ and a diagonal step $\sqrt{2}$). 
The braiding index is then defined as
$\textrm{BI} = \frac{\sum_{e \in E} L_{e}}{L_{r}},$ where $L_{r}$ is the length of the reach measured along the main channel, taken here as the longest path covered by the river~\cite{hong1979study}. It satisfies $\textrm{BI} = 1$ for a single straight channel and increases with both the number of anabranches and their individual sinuosity, so that its deviation from the reference quantifies whether fine channel structure, and not only global radiometry, is preserved.

\end{itemize}

\subsection{Focus on PnP restoration results}


As described in Section~\ref{sec:dataset}, the UNet and DRUNet denoisers used for PnP were trained under three noise-level sampling regimes: a fixed value in $\mathcal{T}_{\text{PnP-fixed}}$, values drawn at random from a discrete set in $\mathcal{T}_{\text{PnP-disc.}}$, and values drawn at random from a continuous distribution in $\mathcal{T}_{\text{PnP-cont.}}$.

Among the denoisers obtained, we have selected only six, as the training schemes that proved to be useable differ from one architecture to another.
For instance, we do not report the results of PnP-UNet trained with a fixed noise level, as they 
produced incoherent reconstructions at inference time.
The same failure is observed for PnP-DRUNet trained with a uniform noise distribution.
Moreover, for the fixed-noise PnP-DRUNet, the restoration quality depends strongly on the standard deviation retained.

The reconstruction obtained from those six configuration are visible in Figure~\ref{fig:pnp-unfolded}, where we considered two synthetic observation from {$\mathcal{V}_{\text{Lhasa-L2S-synth}}$ dataset. These images were chosen to illustrate the variety of land-use patterns at this site. The reconstructions were obtained after 250 iterations, which is sufficient to ensure PSNR convergence and a stable reconstruction.
To maximise the evaluation metrics from the reconstruction, it was also necessary to choose two hyper-parameters namely the regularization parameter $\lambda$ and the step-size $\gamma$ selected for inference and which may vary slightly in order to achieve the best possible reconstructions depending on the considered model.

The results are discussed below:

\begin{itemize}

\item \textbf{UNet discrete:} 
This configuration achieves the best PSNR on both scenes among the three training strategies, together with the water metrics closest to those of the reference. The rural area is well restored, whereas the urban one remains poorly detailed.

\item \textbf{UNet continuous $\mathcal{N}$:} 
Pronounced artefacts appear in the countryside scene, particularly in the red and NIR bands. They are visually identifiable as a green patch in the RGB composite and as a missing region in the corresponding NDWI map. In the urban area, however, this configuration recovers more fine structures than the other two.

\item \textbf{UNet continuous $\mathcal{U}([0,0.2])$:} The rural area is again well restored, but the urban structures are still largely absent.

\item \textbf{DRUNet fixed ($\widetilde{\sigma} = 0.1$)}: This configuration gives the best scores of the whole comparison, both in PSNR/SSIM and in water metrics, and yields markedly sharper and more faithful restorations in the two land-cover types. When $\widetilde{\sigma}$ is set too low (here $0.015$, $0.03$, and $0.045$), the urban was still correctly restored, but the countryside tends towards uniform pixel values, giving an over-smoothed appearance over large regions.

\item \textbf{DRUNet discrete:} The restorations are visually comparable to those of the other two models, but the metrics remain below those of the fixed-$\sigma$.

\item \textbf{DRUNet continuous $\mathcal{N}$:} this configuration largely overestimates the water metrics. The RGB restorations appear visually plausible, yet their PSNR and SSIM drop with respect to the other DRUNet regimes.

\end{itemize}

In all three Unet regimes displayed in Figure~\ref{fig:pnp-unfolded}, the countryside scene, with its more homogeneous land cover, is consistently better restored than the urban scene, which contains a much higher density of fine details. Visually, the DRUNet denoiser clearly outperforms the UNet in all three regimes. The gap is most visible in the urban area, where DRUNet recovers the fine structures of the built environment.

\providecommand{\recblank}{\phantom{\recimg{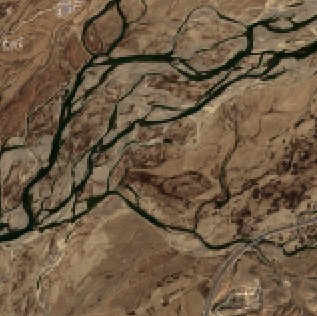}}}
\newlength{\recH}

\begin{figure*}[ht]
	\centering
	\setlength{\tabcolsep}{1.5pt}
	\renewcommand{\arraystretch}{1.1}
	\settoheight{\recH}{\recimg{./images/rec/ref_obs/ref_273.jpg}}%
    \resizebox{\textwidth}{!}{
	\tcbox[recgroup=grpref, title=Sentinel]{%
	\begin{tabular}[t]{c}
		{\footnotesize }\\[-2pt]
		{\footnotesize Sentinel}\\[-1pt]
		\hline\noalign{\smallskip}
		\recimg{./images/rec/ref_obs/ref_273.jpg}\\ \recvoid\\
		\recimg{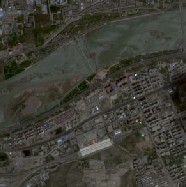}\\ \recvoid\\
		\noalign{\smallskip}\hline\noalign{\smallskip}
		\recimg{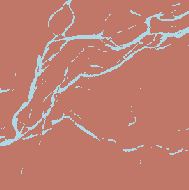}\\
        \recscorewat{63500}{4.66}\\
		\recimg{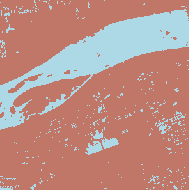}\\
        \recscorewat{195460}{7.88}\\
	\end{tabular}}%
	\hspace{3pt}%
	\tcbox[recgroup=obssyn, title=Synt. Landsat]{%
	\begin{tabular}[t]{c}
    \vspace*{-4.7pt}
		{\footnotesize }\\
		{\footnotesize }\\
		\hline\noalign{\smallskip}
		\recimg{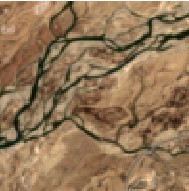}\\ \recvoid\\
		\recimg{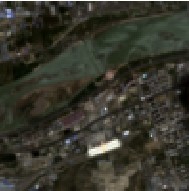}\\ \recvoid\\
		\noalign{\smallskip}\hline\noalign{\smallskip}
		\recimg{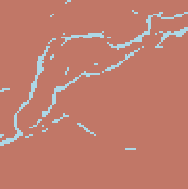}\\
        \recscoredwat{-58\,730}{+0.36}\\
		\recimg{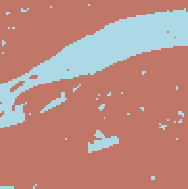}\\
        \recscoredwat{-175\,350}{-5.59}\\
	\end{tabular}}%

	\hspace{3pt}%
	\tcbox[recgroup=calib, title=FB-PnP on $\mathcal{V}_{\text{Lhasa-L2S-synth}}$ ]{%
	\begin{tabular}[t]{ccc@{\hspace{6pt}}ccc}
		\multicolumn{3}{c}{\footnotesize\textbf{UNet}} & \multicolumn{3}{c}{\footnotesize\textbf{DRUNet}}\\[-2pt]
		{\footnotesize disc.} & {\footnotesize cont.-Gaussian} & {\footnotesize cont.-Uniform} &
		{\footnotesize fixed} & {\footnotesize disc.} & {\footnotesize cont.-Gaussian}\\[-1pt]
		\hline\noalign{\smallskip}
		\recimg{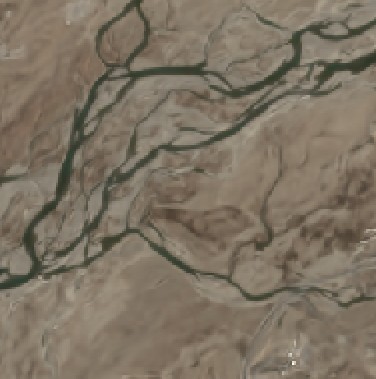} &
		\recimg{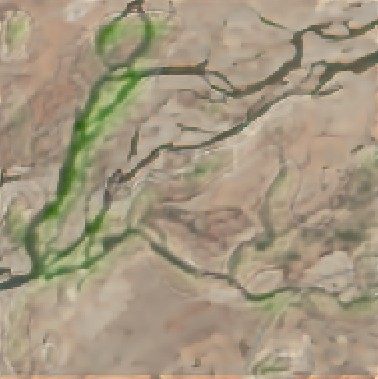} &
		\recimg{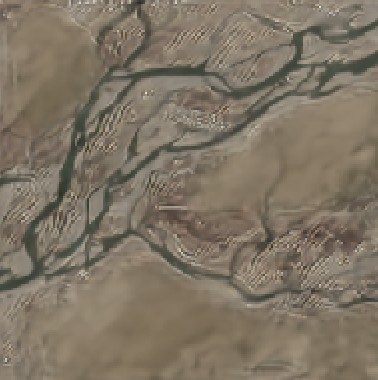} &
		\recimg{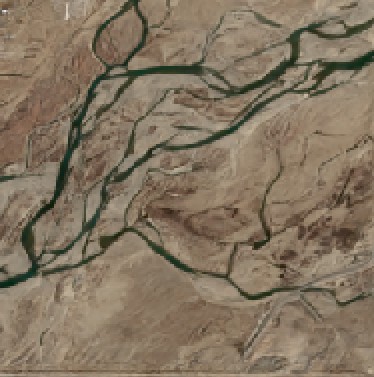} &
		\recimg{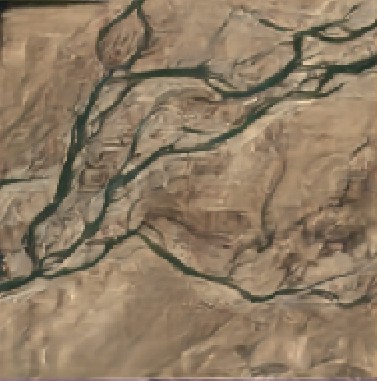} &
		\recimg{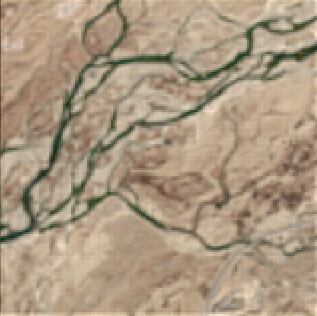}\\
	\recscore{\bstbest{27.2}}{\bstbest{0.74}} & \recscore{22.9}{0.61} & \recscore{\secbest{25.5}}{\secbest{0.71}} &
		\recscore{23.7}{0.59} & \recscore{21.4}{0.51} & \recscore{18.2}{0.42}\\
		\recimg{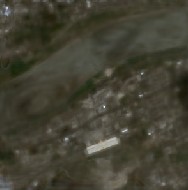} &
		\recimg{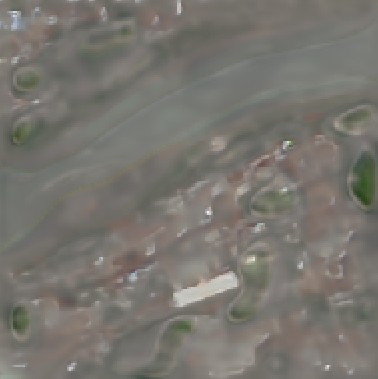} &
		\recimg{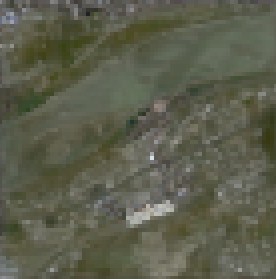} &
		\recimg{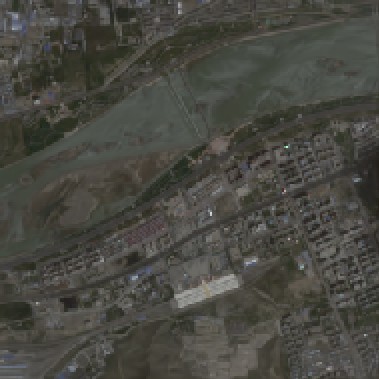} &
		\recimg{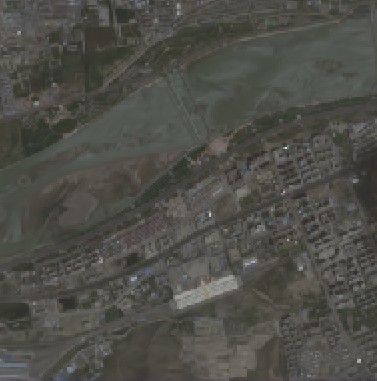} &
		\recimg{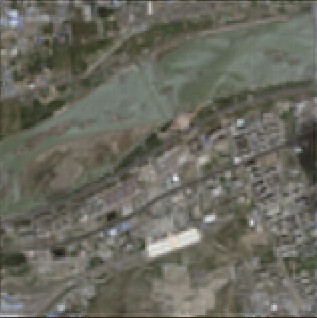}\\
		\recscore{32.5}{0.76} & \recscore{30.4}{0.75} & \recscore{31.4}{0.74} &
		\recscore{\bstbest{41.0}}{\bstbest{0.97}} & \recscore{\secbest{39.4}}{\secbest{0.96}} & \recscore{28.7}{0.64}\\
		\noalign{\smallskip}\hline\noalign{\smallskip}
		\recimg{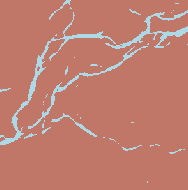} &
		\recimg{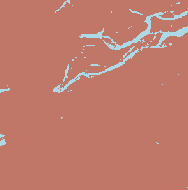} &
		\recimg{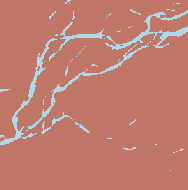} &
		\recimg{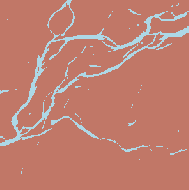} &
		\recimg{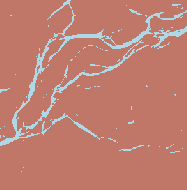} &
		\recimg{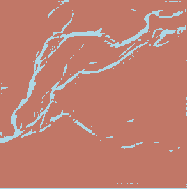}\\
		\recscoredwat{-8\,450}{\bstbest{-0.02}} & \recscoredwat{-27\,710}{+6.94} & \recscoredwat{-8\,440}{-1.55} &
		\recscoredwat{\bstbest{-20}}{-0.74} & \recscoredwat{-2\,590}{\secbest{-0.96}} & \recscoredwat{\secbest{+4\,290}}{+84.99}\\
		\recimg{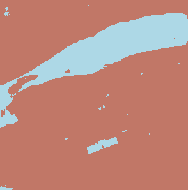} &
		\recimg{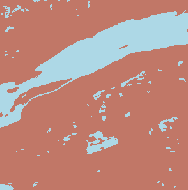} &
		\recimg{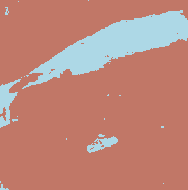} &
		\recimg{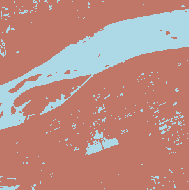} &
		\recimg{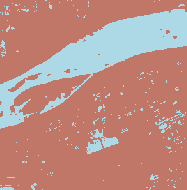} &
		\recimg{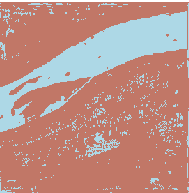}\\
		\recscoredwat{-49\,760}{-5.43} & \recscoredwat{-20\,610}{-3.87} & \recscoredwat{-58\,640}{-3.42} &
		\recscoredwat{\secbest{-500}}{\bstbest{-1.53}} & \recscoredwat{\bstbest{-210}}{\secbest{-2.89}} & \recscoredwat{+15\,160}{+4.96}
	\end{tabular}}%
	\hspace{3pt}%
	\tcbox[recgroup=calib, title=FB-unfolded  on $\mathcal{V}_{\text{Lhasa-L2S-synth}}$]{%
	\begin{tabular}[t]{cc}
		{\footnotesize }\\[-11pt]
		{\footnotesize UNet} & {\footnotesize DRUNet}\\[-1pt]
		\hline\noalign{\smallskip}
		\recimg{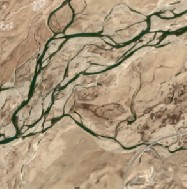} &
		\recimg{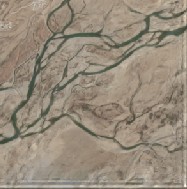}\\
		\recscore{\textbf{31.1}}{\textbf{0.88}} & \recscore{23.2}{0.70}\\
		\recimg{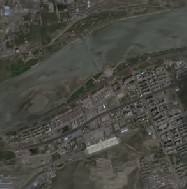} &
		\recimg{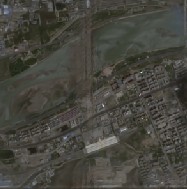}\\
		\recscore{\textbf{39.9}}{0.94} & \recscore{36.8}{\textbf{0.99}}\\
		\noalign{\smallskip}\hline\noalign{\smallskip}
		\recimg{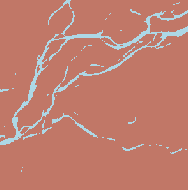} &
		\recimg{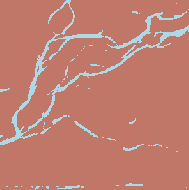}\\
		\recscoredwat{-6\,920}{-1.26} & \recscoredwat{\textbf{+2\,440}}{\textbf{+0.14}}\\
		\recimg{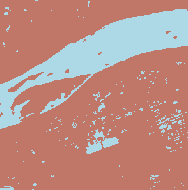} &
		\recimg{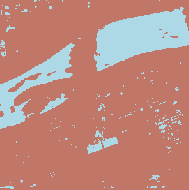}\\
		\recscoredwat{-4\,410}{-3.68} & \recscoredwat{-28\,800}{-0.44}\\
	\end{tabular}}%
	\hspace{1pt}%
	\tcbox[recgroup=grpreal, enhanced, title={\phantom{Xg}},
	       opacityframe=0, opacityback=0, opacitybacktitle=0]{%
	\begin{tabular}[t]{c}
		{\footnotesize }\\[-2pt]
		{\footnotesize }\\[-1pt]
		\noalign{\vskip\arrayrulewidth}\noalign{\smallskip}
		\vphantom{\recimg{./images/rec/ref_obs/ref_273.jpg}}\\
		\vphantom{\recscore{00.00}{0.00}}\\
		\vphantom{\recscore{00.00}{0.00}}\\
		\noalign{\smallskip}\noalign{\vskip\arrayrulewidth}\noalign{\smallskip}
		\vphantom{\recimg{./images/rec/ref_obs/ref_273.jpg}}\\
		\vphantom{\recscoredwat{-000\,000}{+00.00}}\\
		\vphantom{\recscoredwat{-000\,000}{+00.00}}\\
	\end{tabular}}%
    }
	\caption{Reconstructions of two synthetic observations from $\mathcal{V}_{\text{Lhasa-L2S-synth}}$ dataset with FB-PnP (UNet and DRUNet denoisers) and FB-unfolded networks are displayed in line 1 and 2 and their Thresholded-NDWI counterpart (i.e. $\mathrm{t}^{\mathrm{NDWI}}$) in line 3 and 4. For each scene, the RGB composite is reported with its PSNR (dB) and SSIM. For the NDWI maps, the first column reports the reference values of the open-water surface area WA [m$^2$] and of the braiding index BI (dimensionless). The values reported for observation and the reconstructions are the signed differences with respect to these references, $\Delta = \text{value} - \text{reference}$, so that values closer to zero indicate better agreement. Per metric, \bstbest{light blue bold} marks the best score and \secbest{blue bold} the second best for PnP and in \textbf{black bold} for unfolded.}
	\label{fig:pnp-unfolded}
\end{figure*}

In the remainder of the paper, we focus on the \textbf{PnP DRUNet fixed} configuration, which outperforms all other approaches considered for every metric.

\begin{figure*}[h]
	\centering
	\setlength{\tabcolsep}{1.5pt}
	\renewcommand{\arraystretch}{1.1}
	\resizebox{\textwidth}{!}{
	\tcbox[recgroup=grpref, title=Target]{%
	\begin{tabular}[t]{c}
		{\footnotesize Sentinel}\\[2.5pt]
		\recimg{./images/rec/ref_obs/ref_273.jpg}\\ \recvoid\\ \noalign{\medskip}
		\recimg{./images/rec/ref_obs/ref_160.jpg}\\ \recvoid\\ \noalign{\medskip}
		\recimg{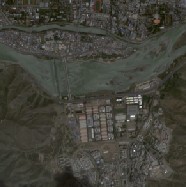}\\  \recvoid\\ \noalign{\medskip}
		\recimg{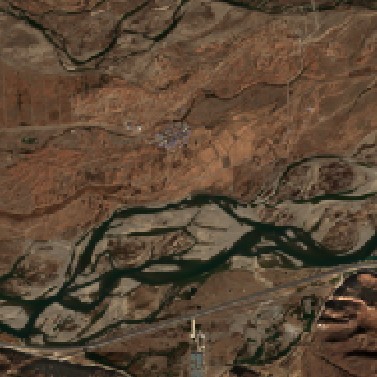}\\ \recvoid\\
	\end{tabular}}%
	\hspace{3pt}%
	\tcbox[recgroup=obssyn, title=Observation]{%
	\begin{tabular}[t]{c}
		{\footnotesize Synt. Landsat}\\[2.5pt]
		\recimg{images/rec/ref_obs/synth_obs/obs_273.jpg}\\ \recvoid\\ \noalign{\medskip}
		\recimg{images/rec/ref_obs/synth_obs/obs_160.jpg}\\ \recvoid\\ \noalign{\medskip}
		\recimg{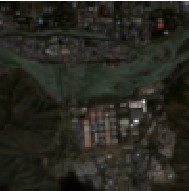}\\  \recvoid\\ \noalign{\medskip}
		\recimg{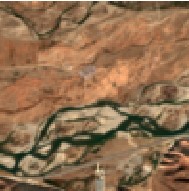}\\ \recvoid\\
	\end{tabular}}%
	\hspace{3pt}%
	\tcbox[recgroup=grpsyn, title=Reconstruction on $\mathcal{V}_{\text{Lhasa-L2S-synth}}$]{%
	\begin{tabular}[t]{cccc}
		{\footnotesize FB-PnP} & {\footnotesize Unfolded-UNet} & {\footnotesize SwinIR} & {\footnotesize RDN}\\
		\recimg{images/rec/PnP/PnP_DRUNet_fn/DRUNet_im273_sigma01_psnr23p69_0p59.jpg}
		& \recimg{images/rec/unfolded/UNet/im273/im273_normalized_psnr31p13_0p88.jpg}
		& \recimg{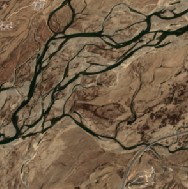}
		& \recimg{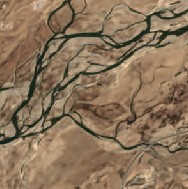}\\
		\recscore{23.7}{0.59} & \recscore{\secbest{31.1}}{\secbest{0.88}} & \recscore{\bstbest{35.1}}{\bstbest{0.94}} & \recscore{30.7}{0.83}\\
		\noalign{\medskip}
		\recimg{images/rec/PnP/PnP_DRUNet_fn/DRUNet_im160_sigma01_psnr41p04_0p97.jpg}
		& \recimg{images/rec/unfolded/UNet/im160/im160_normalized_psnr39p87_0p94.jpg}
		& \recimg{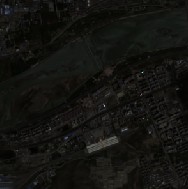}
		& \recimg{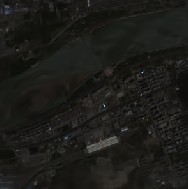}\\
		\recscore{\secbest{41.0}}{\secbest{0.97}} & \recscore{39.9}{0.94} & \recscore{\bstbest{44.1}}{\bstbest{0.98}} & \recscore{38.7}{0.92}\\
		\noalign{\medskip}
		\recimg{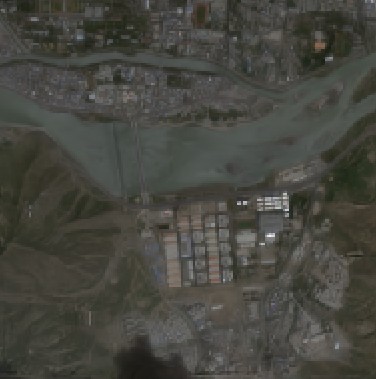}
		& \recimg{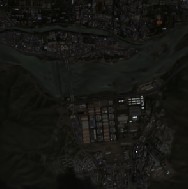}
		& \recimg{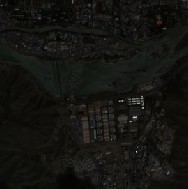}
		& \recimg{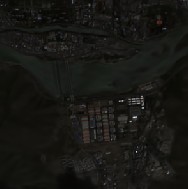}\\
		\recscore{\secbest{41.7}}{\secbest{0.97}} & \recscore{40.8}{0.96} & \recscore{\bstbest{45}}{\bstbest{0.98}} & \recscore{39.4}{0.93}\\
		\noalign{\medskip}
		\recimg{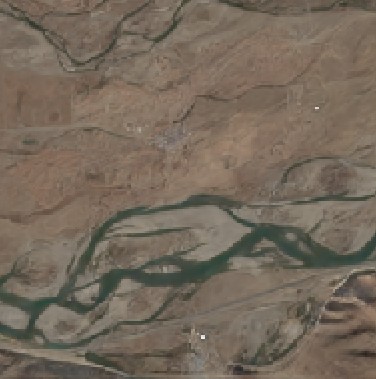}
		& \recimg{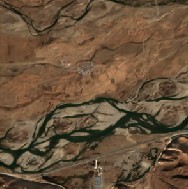}
		& \recimg{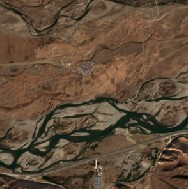}
		& \recimg{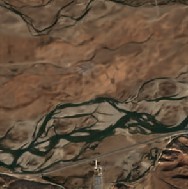}\\
		\recscore{26.9}{0.68} & \recscore{\secbest{33.3}}{\secbest{0.89}} & \recscore{\bstbest{37.9}}{\bstbest{0.96}} & \recscore{32.9}{0.83}\\
	\end{tabular}}%
	\hspace{3pt}%
	\tcbox[recgroup=obsreal, title=Observation]{%
	\begin{tabular}[t]{c}
        \vspace*{5pt}
		{\footnotesize Landsat}\\
		\recimg{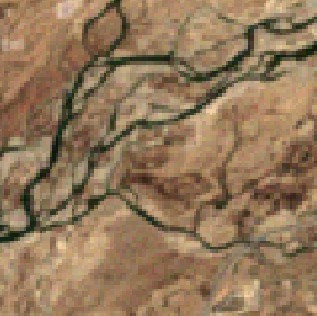}\\ \recvoid\\ \noalign{\medskip}
		\recimg{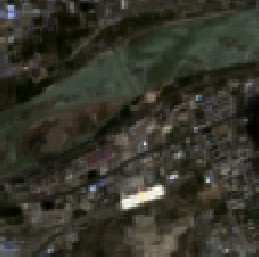}\\ \recvoid\\ \noalign{\medskip}
		\recimg{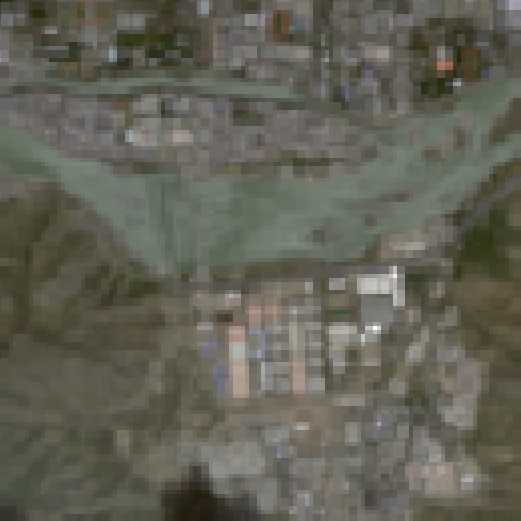}\\  \recvoid\\ \noalign{\medskip}
		\recimg{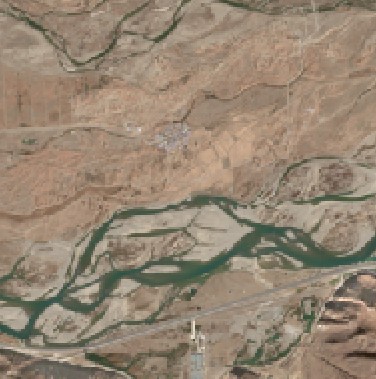}\\ \recvoid\\
	\end{tabular}}%
	\hspace{3pt}%
	\tcbox[recgroup=grpreal, title=Reconstruction on $\mathcal{V}_{\text{Lhasa-L2S}}$]{%
	\begin{tabular}[t]{cccc}
		{\footnotesize FB-PnP} & {\footnotesize Unfolded-UNet} & {\footnotesize SwinIR} & {\footnotesize RDN}\\[2.3pt]
		\recimg{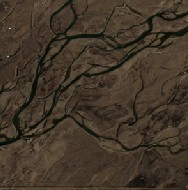}
		& \recimg{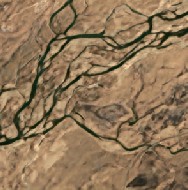}
		& \recimg{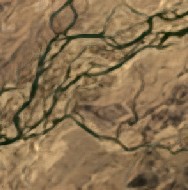}
		& \recimg{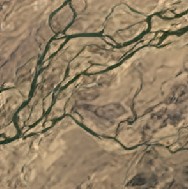}\\
		\recscore{16.0}{0.45} & \recscore{\bstbest{18.2}}{\bstbest{0.68}} & \recscore{\secbest{16.9}}{\secbest{0.61}} & \recscore{16.5}{0.61}\\
		\noalign{\medskip}
		\recimg{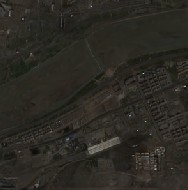}
		& \recimg{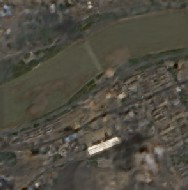}
		& \recimg{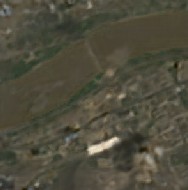}
		& \recimg{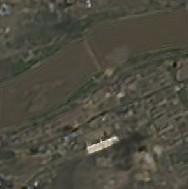}\\
		\recscore{\secbest{10.2}}{0.21} & \recscore{\bstbest{10.4}}{\bstbest{0.25}} & \recscore{10.1}{0.25} & \recscore{10.1}{\secbest{0.23}}\\
		\noalign{\medskip}
		\recimg{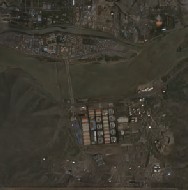}
		& \recimg{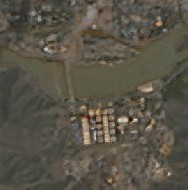}
		& \recimg{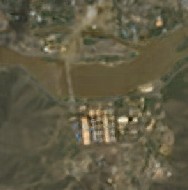}
		& \recimg{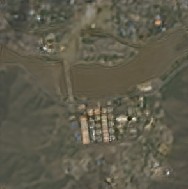}\\
		\recscore{\secbest{9}}{0.19} & \recscore{\bstbest{9.17}}{\secbest{0.23}} & \recscore{8.9}{\bstbest{0.24}} & \recscore{8.9}{0.22}\\
		\noalign{\medskip}
		\recimg{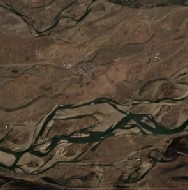}
		& \recimg{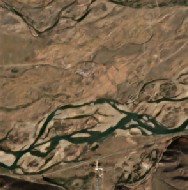}
		& \recimg{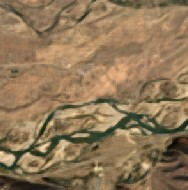}
		& \recimg{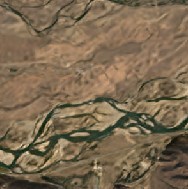}\\
		\recscore{14.7}{0.51} & \recscore{\bstbest{15.8}}{\bstbest{0.71}} & \recscore{\secbest{15.0}}{\secbest{0.65}} & \recscore{14.9}{0.63}\\
	\end{tabular}}%
	}
	\caption{Reconstruction comparison on four scenes. From left to right: Sentinel reference, synthetic observation, reconstructions from the synthetic observation, true Landsat-8/9 observation, and reconstructions from the true observation. Methods: the proposed model-based approaches (FB-PnP and unfolded forward--backward) and the SwinIR and RDN baselines. Per line and per metric, \bstbest{light blue bold} marks the best score and \secbest{blue bold} the second best.}
	\label{fig:rec_all_methods}
\end{figure*}

\begin{figure*}[h]
	\centering
	\setlength{\tabcolsep}{1.5pt}
	\renewcommand{\arraystretch}{1.1}
	\resizebox{\textwidth}{!}{
	\tcbox[recgroup=grpref, title=Target]{%
	\begin{tabular}[t]{c}
		{\footnotesize Sentinel}\\
		\recimg{images/rec/ref_obs/NDWI_im273_ref_WA63500_4p66.png}\\
        \recscorewat{63500}{4.66} \\ \noalign{\medskip}
		\recimg{images/rec/ref_obs/NDWI_im160_ref_WA195460_7p88.png}\\
        \recscorewat{195460}{7.88}\\ \noalign{\medskip}
		\recimg{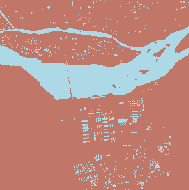}\\
        \recscorewat{165020}{7.35}\\ \noalign{\medskip}
		\recimg{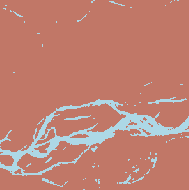}\\
        \recscorewat{83150}{4.60}\\
	\end{tabular}}%
	\hspace{3pt}%
	\tcbox[recgroup=obssyn, title=Observation]{%
	\begin{tabular}[t]{c}
		{\footnotesize Synt.  Landsat}\\[2.5pt]
		\recimg{images/rec/ref_obs/synth_obs/obssynth_273_WA4770_5p02.png}\\
        \recscoredwat{-20\,570}{+0.36}\\ \noalign{\medskip}
		\recimg{images/rec/ref_obs/synth_obs/obssynth_160_WA20110_2p29.png}\\
        \recscoredwat{-14\,470}{-5.59}\\ \noalign{\medskip}
		\recimg{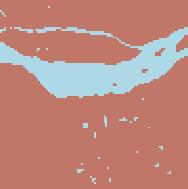}\\
        \recscoredwat{-19\,400}{-5.15}\\ \noalign{\medskip}
		\recimg{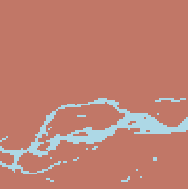}\\
        \recscoredwat{-12\,950}{-2.15}\\
	\end{tabular}}%
	\hspace{3pt}%
	\tcbox[recgroup=grpsyn, title=Reconstruction on $\mathcal{V}_{\text{Lhasa-L2S-synth}}$]{%
	\begin{tabular}[t]{cccc}
		{\footnotesize FB-PnP} & {\footnotesize Unfolded-UNet} & {\footnotesize SwinIR} & {\footnotesize RDN}\\
		\recimg{images/rec/PnP/PnP_DRUNet_fn/NDWI/sigma01/NDWI_im273_WA63480_BI3p92.png}
		& \recimg{images/rec/unfolded/UNet/im273/NDWI_im273_WA56580_BI3p4.png}
		& \recimg{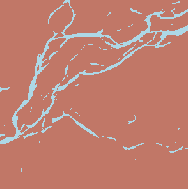}
		& \recimg{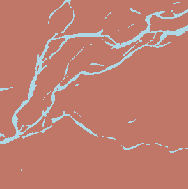}\\
		\recscoredwat{\bstbest{-20}}{-0.74} & \recscoredwat{-6\,920}{-1.26} & \recscoredwat{-15\,700}{\secbest{-0.51}} & \recscoredwat{\secbest{-3\,580}}{\bstbest{+0.47}}\\
		\noalign{\medskip}
		\recimg{images/rec/PnP/PnP_DRUNet_fn/NDWI/sigma01/NDWI_im160_WA194960_BI6p35.png}
		& \recimg{images/rec/unfolded/UNet/im160/NDWI_im160_WA191050_BI4p2.png}
		& \recimg{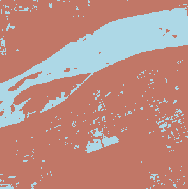}
		& \recimg{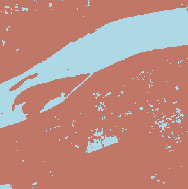}\\
		\recscoredwat{\bstbest{-500}}{\bstbest{-1.53}} & \recscoredwat{-4\,410}{\secbest{-3.68}} & \recscoredwat{-14\,580}{-4.29} & \recscoredwat{\secbest{-2\,500}}{-5.11}\\
		\noalign{\medskip}
		\recimg{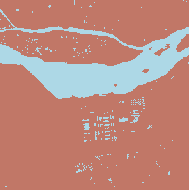}
		& \recimg{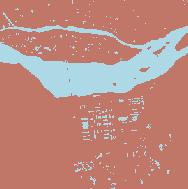}
		& \recimg{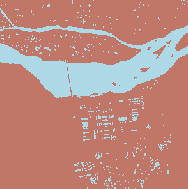}
		& \recimg{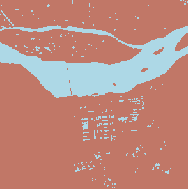}\\
		\recscoredwat{\bstbest{-1\,150}}{\secbest{-4.15}} & \recscoredwat{\secbest{-1\,250}}{-4.79} & \recscoredwat{-17\,000}{-5.04} & \recscoredwat{-5\,110}{\bstbest{-4.08}}\\
		\noalign{\medskip}
		\recimg{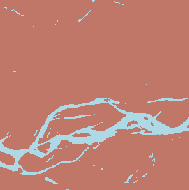}
		& \recimg{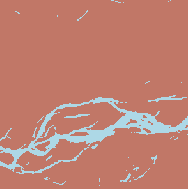}
		& \recimg{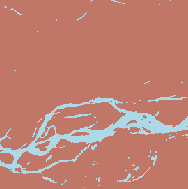}
		& \recimg{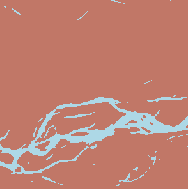}\\
		\recscoredwat{\secbest{-6\,320}}{\bstbest{-1.03}} & \recscoredwat{-8\,740}{\secbest{-1.05}} & \recscoredwat{-11\,860}{-1.66} & \recscoredwat{\bstbest{-4\,540}}{-1.26}\\
	\end{tabular}}%
	\hspace{3pt}%
	\tcbox[recgroup=obsreal, title=Observation]{%
	\begin{tabular}[t]{c}
		{\footnotesize Landsat}\\[3.5pt]
		\recimg{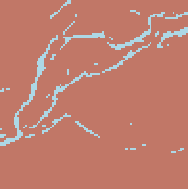}\\
        \recscoredwat{-10\,310}{-0.34}\\ \noalign{\medskip}
		\recimg{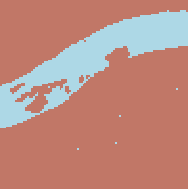}\\
        \recscoredwat{-21\,580}{-5.58}\\ \noalign{\medskip}
		\recimg{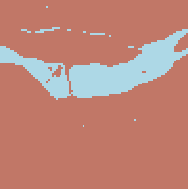}\\
        \recscoredwat{-40\,460}{-4.99}\\ \noalign{\medskip}
		\recimg{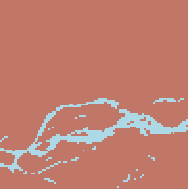}\\
        \recscoredwat{-20\,240}{-2.33}\\
	\end{tabular}}%
	\hspace{3pt}%
	\tcbox[recgroup=grpreal, title=Reconstruction on $\mathcal{V}_{\text{Lhasa-L2S}}$]{%
	\begin{tabular}[t]{cccc}
		{\footnotesize FB-PnP} & {\footnotesize Unfolded-UNet} & {\footnotesize SwinIR} & {\footnotesize RDN}\\[2.5pt]
		\recimg{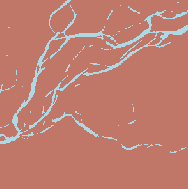}
		& \recimg{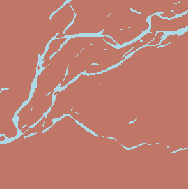}
		& \recimg{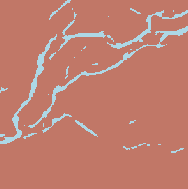}
		& \recimg{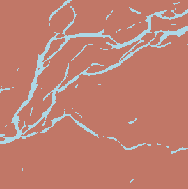}\\
		\recscoredwat{\bstbest{-2\,740}}{-1.98} & \recscoredwat{\secbest{-6\,060}}{\secbest{-1.49}} & \recscoredwat{-7\,400}{\bstbest{+0.21}} & \recscoredwat{+6\,860}{-2.27}\\
		\noalign{\medskip}
		\recimg{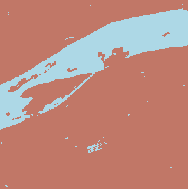}
		& \recimg{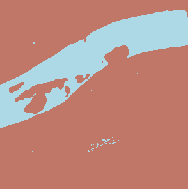}
		& \recimg{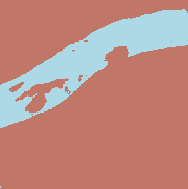}
		& \recimg{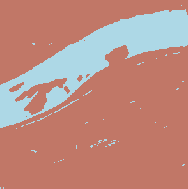}\\
		\recscoredwat{-25\,730}{\bstbest{-3.79}} & \recscoredwat{-22\,130}{-5.46} & \recscoredwat{\secbest{-20\,760}}{-5.11} & \recscoredwat{\bstbest{-14\,980}}{\secbest{-5.04}}\\
		\noalign{\medskip}
		\recimg{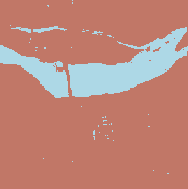}
		& \recimg{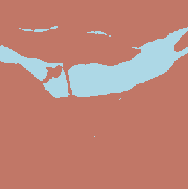}
		& \recimg{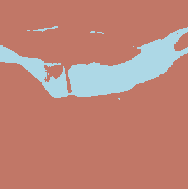}
		& \recimg{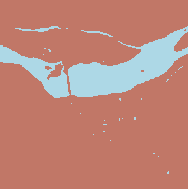}\\
		\recscoredwat{\bstbest{-30\,480}}{\secbest{-3.54}} & \recscoredwat{-44\,710}{-4.36} & \recscoredwat{-46\,620}{-5.53} & \recscoredwat{\secbest{-38\,050}}{\bstbest{-3.08}}\\
		\noalign{\medskip}
		\recimg{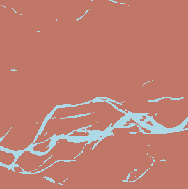}
		& \recimg{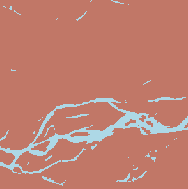}
		& \recimg{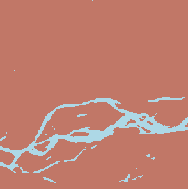}
		& \recimg{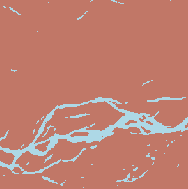}\\
		\recscoredwat{\secbest{-17\,330}}{\bstbest{-0.53}} & \recscoredwat{-17\,370}{\secbest{-1.01}} & \recscoredwat{-20\,670}{-1.22} & \recscoredwat{\bstbest{-10\,710}}{-1.24}\\
	\end{tabular}}%
	}
	\caption{Segmentation of the reconstructions of four scenes. From left to right: NDWI of the Sentinel reference, with the absolute values of the open-water surface area WA [m$^2$] and of the braiding index BI (dimensionless); NDWI of the synthetic observation; segmentations obtained from the synthetic observation; NDWI of the true Landsat-8/9 observation; and segmentations obtained from the true observation. For the observation and reconstruction blocks, the two metrics are reported as signed differences with respect to the reference, $\Delta = \text{value} - \text{reference}$, so that a value closer to zero indicates a better agreement. Per line and per metric, \bstbest{light blue bold} marks the best score and \secbest{blue bold} the second best, computed over the four reconstruction methods only. The WA of the two observation blocks has been rescaled by a factor of 9 to account for the Landsat ground sampling distance of 30~m against 10~m for Sentinel-2.}
	\label{fig:seg_all_methods}
\end{figure*}

\subsection{Focus on Unfolded restoration results} 
As for the PnP approach, we compared the UNet and DRUNet architectures within the unfolded approach. Figure~\ref{fig:pnp-unfolded} shows that, on both synthetic observations, FB-unfolded with the UNet achieves a higher PSNR than with the DRUNet. The visual quality is consistent with these scores: the spectral dynamics are well recovered both in the RGB bands and in the NIR band, the latter being assessed through the NDWI.
With the DRUNet, the rural scene is visually well reconstructed, whereas the urban scene is strongly degraded: a large linear artefact appears in all bands, and is most visible in the RGB composite.

Given its higher PSNR and its stable reconstructions across both scenes, the \textbf{FB-unfolded UNet} is retained for the remainder of the paper.

\subsection{Comparisons between all reconstruction strategies}
We finally compare the two proposed model-based approaches FB-PnP (for \textit{FB-PnP DRUNet fixed}) and FB-unfolded  (for \textit{FB-unfolded UNet}) against two state-of-the-art black-box learned restoration networks, SwinIR and RDN, both on synthetic {$\mathcal{V}_{\text{Lhasa-L2S-synth}}$ and real observations {$\mathcal{V}_{\text{Lhasa-L2S}}$}. The results are displayed in Figure~\ref{fig:rec_all_methods} and Figure~\ref{fig:seg_all_methods}.

\noindent \textbf{Inference on synthetic observations --} On PSNR and SSIM, SwinIR dominates the comparison, with averages of $40.5$~dB and $0.96$ over the four scenes, ahead of the FB-unfolded network ($36.3$~dB, $0.92$), RDN ($35.4$~dB, $0.878$), and FB-PnP ($33.3$~dB, $0.80$). 
The ordering is stable from one scene to the other. 
FB-PnP is also the most irregular approach, with $41.0$ and $41.7$~dB on urban area against $23.7$ and $26.9$~dB on rural one, a spread of $18.0$~dB that reflects the sensitivity of the fixed-$\sigma$ regime to the land cover of the scene, whereas the three learned networks vary by less than $11$~dB over the same scenes.

The water metrics reverse this ranking entirely. 
On the open-water surface area, FB-PnP is the closest to the Sentinel reference with a mean relative error of $2.2\%$, ahead of RDN ($3.9\%$), the unfolded network ($6.1\%$) and SwinIR ($14.2\%$).
Considering the braiding index, the mean absolute error is again the lowest for FB-PnP ($1.86$), ahead of the unfolded network ($2.70$), RDN ($2.73$) and SwinIR ($2.88$). 
The rural scenes underline this inversion the most: FB-PnP obtains the worst PSNR of the four methods, $23.69$~dB against $35.12$~dB for SwinIR, yet recovers the reference water surface to within $0.03\%$ ($63\,480 \quad m²$ against $63\,500 \quad m²$), while SwinIR loses a quarter of it ($47\,800$, that is $-24.7\%$). 
In other words, the method that best reproduces the radiometry of the reference is the one that least preserves its hydromorphological content, in terms of water channel network.

\noindent \textbf{Inference on real observations --} The FB-unfolded approach achieves the highest PSNR on all four scenes, with an average of $13.4$~dB against $12.7$~dB for SwinIR, $12.6$~dB for RDN and $12.5$~dB for FB-PnP, and the highest average SSIM ($0.47$ against $0.44$, $0.42$ and $0.34$ respectively). 
The gain is most pronounced on the non-urban scenes, where it reaches $+1.3$~dB and $+0.8$~dB in PSNR and $+0.07$ and $+0.06$ in SSIM over SwinIR, the best of the two baselines. 
SwinIR and RDN yield very close results, while FB-PnP remains competitive in PSNR but systematically obtains the lowest SSIM. 
In the urban area, second and third lines, all methods remain within a narrow range about $0.3$~dB in PSNR and $0.05$ in SSIM.

The two water metrics give again a different ordering. 
On the open-water surface area, RDN is the closest to the Sentinel reference with a mean relative error of $13.6\%$, followed by FB-PnP ($14.2\%$), the unfolded network ($17.2\%$) and SwinIR ($18.9\%$).
For the braiding index, FB-PnP is the most faithful with a mean absolute error of $2.46$, ahead of RDN ($2.91$), SwinIR ($3.02$) and the unfolded network ($3.08$). 
The method that performs best in PSNR and SSIM yields slightly larger deviations on both hydrological variables, although these differences remain small compared with those on reconstruction metrics. This is consistent with the decoupling observed on synthetic data: higher reconstruction performance does not necessarily imply better preservation of hydromorphological content.}


\noindent\textbf{Summary --} On real observations, which depart from the training distribution, the unfolded network clearly outperforms all other methods: it achieves the highest PSNR and the most visually convincing reconstructions across all scenes, at a reasonable computational cost (cf. Table~\ref{tab:training}). Regarding the extraction of the braiding index and the water area, all methods perform reasonably well, with an advantage for FB-PnP and RDN. This suggests that the preservation of hydromorphological descriptors of the water channel network should be explicitly integrated into the design of the unfolded model, which we leave for future work.


\subsection{Statistical Significance on $\mathcal{V}_{\text{Lhasa-L2S}}$ }
\label{sec:stat}
Whereas the previous subsection was restricted to four validation images, the analysis is here extended to the whole validation set.  To test whether the methods differ in performance, we apply the Friedman test \cite{friedman1937use} to the per-image ranks of each metric. 
For every image, the methods are ranked from best to worst. 
The test then asks whether some methods tend to receive better ranks than others across the whole set. 
Its null hypothesis is that all methods perform equally, meaning that none is systematically better or worse. 
The alternative hypothesis is the opposite: at least one method is systematically better or worse than the rest. 
The test summarises the observed rank differences in a single statistic:
\begin{equation}
    \chi^2_F = \frac{12}{n\,k(k+1)} \sum_{j=1}^{k} (\sum_{i=1}^N r_{ij})^{\,2} \;-\; 3\,n(k+1)
\end{equation}
where $n=690$ is the number of images, $k=4$ the number of methods compared, and $r_{ij}\in \{1,2,3,4\}$  denote the rank of method $j$ on image $i$ (1 for the best method). The larger $\chi^2_F$, and the stronger the evidence against the null hypothesis. 

Table~\ref{tab:pooled_friedman} summarizes the behaviour of the four methods.  In terms of reconstruction quality (PSNR, SSIM), FB-Unfolded and SwinIR clearly outperform the other two methods: FB-Unfolded reaches the highest PSNR (16.40\,dB), closely followed by SwinIR (16.34\,dB), while SwinIR obtains the best SSIM (0.402) ahead of FB-Unfolded (0.374). The ranking is however different, and partly reversed, for the hydromorphological indicators. All methods underestimate the braiding index of the Sentinel reference, but FB-PnP and RDN performs better. The Friedman test rejects the null hypothesis of equal performance for all four metrics; the much larger statistics are obtained for PSNR and SSIM, which indicates that the methods are ranked far more consistently across reconstruction metrics than on hydromorphological indicators. The Friedman statistics are very large for both reconstruction metrics, confirming that these differences are highly unlikely to arise by chance.

\begin{table*}[t]
\centering
\setlength{\tabcolsep}{8pt}
\renewcommand{\arraystretch}{1.15}
\begin{tabular}{l c c c c}
\toprule
\textbf{Method} & \textbf{PSNR} (dB)$\uparrow$ & \textbf{SSIM}$\uparrow$
 & \textbf{Braiding index} & \textbf{Open-water area} (m$^2$)\\
 & & & $\Delta$ to ref. & $\Delta$ to ref. \\
\midrule
Sentinel reference & --- & --- & 7.30\,$\pm$\,3.94 & 246413\,$\pm$\,151731 \\
\midrule
FB-PnP-DRUNet    & 15.56\,$\pm$\,3.36 & 0.343\,$\pm$\,0.094 & \bstbest{$-$1.31} & \secbest{$-$963} \\
FB-Unfolded-UNet & \bstbest{16.40}\,$\pm$\,3.46 & \secbest{0.374}\,$\pm$\,0.100 & $-2.62$ & $+1109$ \\
SwinIR           & \secbest{16.34}\,$\pm$\,3.74 & \bstbest{0.402}\,$\pm$\,0.097 & $-2.46$ & \bstbest{$+$650} \\
RDN              & 15.61\,$\pm$\,3.42 & 0.338\,$\pm$\,0.098 & \secbest{$-$1.66} & $+13132$ \\
\noalign{\hrule height \arrayrulewidth}
\textbf{Friedman $\chi^2_F$} & 1520.2 & 1833.6 & 780.7 & 68.9 \\
\bottomrule
\end{tabular}
\caption{Mean\,$\pm$\,std over the set; $\uparrow$/$\downarrow$ = higher/lower is better.
For the water metrics, the first row gives the absolute values measured on the Sentinel reference, and each method is reported as $\Delta$, the signed difference between its mean and that of the reference.
\bstbest{Light blue} = best per column and \secbest{dark blue} = second best, the $\Delta$ columns being judged on the smallest absolute value.
The Friedman test rejects the null hypothesis of equal performance for all four metrics.}
\label{tab:pooled_friedman}
\end{table*}

\section{Conclusion}
\label{sec:conclusion}

This work introduced, to the best of our knowledge, the first dataset
dedicated to the super-resolution of multispectral satellite images for river
analysis. Its construction relies on the period during which Landsat-8/9
and Sentinel-2 acquired the same areas: the higher-resolution Sentinel-2
images provide a reference against which the enhancement of Landsat
observations can be assessed, and this strategy can be extended to build
larger training and evaluation sets for increasing the resolution of the
long Landsat archive.

On this dataset, we developed a plug-and-play (PnP) approach and an unfolded forward-backward network, and compared them against state-of-the-art black-box NN restoration methods. Evaluating the four methods in three complementary ways (on synthetic observations, where the degradation operator is known exactly; on real Landsat-8/9 observations, where it is not; and through a statistical test over the whole real validation dataset) allowed us to separate their respective strengths and weaknesses.

For the reconstruction of the RGB components, FB-Unfolded is the most accurate method, leading both the PSNR and the SSIM comparisons on every real scene. Combined with its fast inference, it offers the best compromise between computational cost and visual reconstruction quality.

The conclusion is more nuanced for river descriptors. For the
segmentation of the water surface derived from the NDWI index, PnP and RDN yield better scores, although the statistical test does not show a clear advantage of these methods over the FB-unfolded network. This suggests that the unfolded architecture remains a promising candidate, provided that it is designed with segmentation as an explicit objective rather than image fidelity alone, for instance following the approach of~\cite{le2025embedding}. Such task-oriented unfolded networks, trained on larger Landsat/Sentinel-2 datasets, constitute a natural direction for future work.

The study of high spatial and spectral satellite image time series remains a challenging question \cite{pelletier2019temporal,ruwurm2020,gomez2016}. In the present contribution we aim to propose a framework that enhances the resolution of past images to build a dataset covering a much longer period, thereby densifying the time series and makes multi-decadal change analysis possible.


\section*{Acknowledgment}
B.~Belletti's contribution is framed within the CPJ Hydrosyst\`emes supported by the French National Research Agency (ANR-23-CPJ1-0148-01).
This work was supported by the French National Research Agency (ANR) through the GloUrb project, ANR-22-CE03-0005, and the DeepScale project, ANR-26-CE48-5044-01.
We thank the Centre Blaise Pascal (CBP) of the ENS de Lyon for granting us access to a high-performance computer to carry out the numerical experiments.

\bibliographystyle{IEEEtran}
\bibliography{biblio}

\end{document}